\documentclass[10pt,journal,final,twocolumn]{IEEEtran}
\usepackage{graphicx,times,amsmath,amssymb,cite,cases,mathrsfs,booktabs,multirow,psfrag,subfigure,url,stfloats}
\usepackage{algorithm,algorithmic,amsthm,ulem,latexsym,bm,bbm}
\usepackage[colorlinks,linkcolor=red,anchorcolor=blue,citecolor=red]{hyperref}
\DeclareMathOperator*{\argmax}{argmax}

\newcommand{\etal}{et al.}

\begin{document}
%
\title{Learning Target Domain Specific Classifier for Partial Domain Adaptation}

\author{Chuan-Xian Ren, Pengfei Ge, Peiyi Yang, Shuicheng Yan
\thanks{C.X. Ren, P.F. Ge, and P.Y. Yang are with the School of Mathematics, Sun Yat-Sen University, Guangzhou, China. C.X. Ren is the corresponding author. (email: rchuanx@mail.sysu.edu.cn)}
\thanks{S.C. Yan is with the Yitu Technology, Beijing, China.}
\thanks{This work is supported in part by the National Natural Science Foundation of China under Grants 61976229, 61906046, 61572536, 11631015, and U1611265.}}

\date{}
\IEEEcompsoctitleabstractindextext{%
\begin{abstract}%
Unsupervised domain adaptation~(UDA) aims at reducing the distribution discrepancy when transferring knowledge from a labeled source domain to an unlabeled target domain. Previous UDA methods assume that the source and target domains share an identical label space, which is unrealistic in practice since the label information of the target domain is agnostic. This paper focuses on a more realistic UDA scenario, i.e. partial domain adaptation (PDA), where the target label space is subsumed to the source label space. In the PDA scenario, the source outliers that are absent in the target domain may be wrongly matched to the target domain (technically named \textit{negative transfer}), leading to performance degradation of UDA methods.
This paper proposes a novel Target Domain Specific Classifier Learning-based Domain Adaptation (TSCDA) method. TSCDA presents a soft-weighed maximum mean discrepancy criterion to partially align feature distributions and alleviate negative transfer. Also, it learns a target-specific classifier for the target domain  with pseudo-labels and multiple auxiliary classifiers, to further address classifier shift. A module named Peers Assisted Learning is used to minimize the prediction difference between multiple target-specific classifiers, which makes the classifiers more discriminant for the target domain. Extensive experiments conducted on three PDA benchmark datasets show that TSCDA outperforms other state-of-the-art methods with a large margin, e.g. $4\%$ and $5.6\%$ averagely on \textit{Office-31} and \textit{Office-Home}, respectively.
\end{abstract}

}

\maketitle \IEEEdisplaynotcompsoctitleabstractindextext \IEEEpeerreviewmaketitle

\section{Introduction}\label{sec:introduction}

\IEEEPARstart{D}{eep} Convolutional Neural Networks (CNNs) have made remarkable advances in a variety of machine learning tasks such as image classification~\cite{GoogleNet-v1,lin2019unsupervised,ResNet}, clustering analysis~\cite{ge2019dual} and object detection~\cite{R-CNN,R-FCN}. Unfortunately, a model trained in the source domain (training set) usually suffers severe performance degradation in a target domain (test set) with a distribution difference from the source domain~\cite{fang2018flexible}. This difference is termed \textit{domain shift}, which is the bottleneck of many practical cross-domain applications. To solve it, UDA methods~\cite{ADDA,DAN,SAN,DAsurvey,ren2018generalized} try to transfer knowledge from a label-rich source domain to an unlabeled target domain. However, these methods need to assume the source domain shares the same label space with the target domain, i.e., $\mathcal{Y}^s=\mathcal{Y}^t$, where $\mathcal{Y}^s $ and $\mathcal{Y}^t$ denote the label spaces of source and target domains, respectively. This assumption is hard to satisfy since in the real world, it is almost unrealistic to find a source domain that has the identical label space with the target domain considering the label information of the target domain is agnostic. In this paper, we consider a more realistic task setting, i.e., PDA~\cite{PADA,IWAN,AFN}, where the target label space is a subset of the source label space, i.e., $\mathcal{Y}^t\subset\mathcal{Y}^s$. Specifically, we divide the source domain into a \textit{source-shared-domain} where the subjects are covered by both domains, and a \textit{source-outlier-domain} where the subjects only exist in the source domain. 

\begin{figure*}[ht]
\subfigure[]{\label{fig:diagram1_a}
\begin{minipage}[c]{0.33\textwidth}
\centering \scalebox{0.5}{
\includegraphics{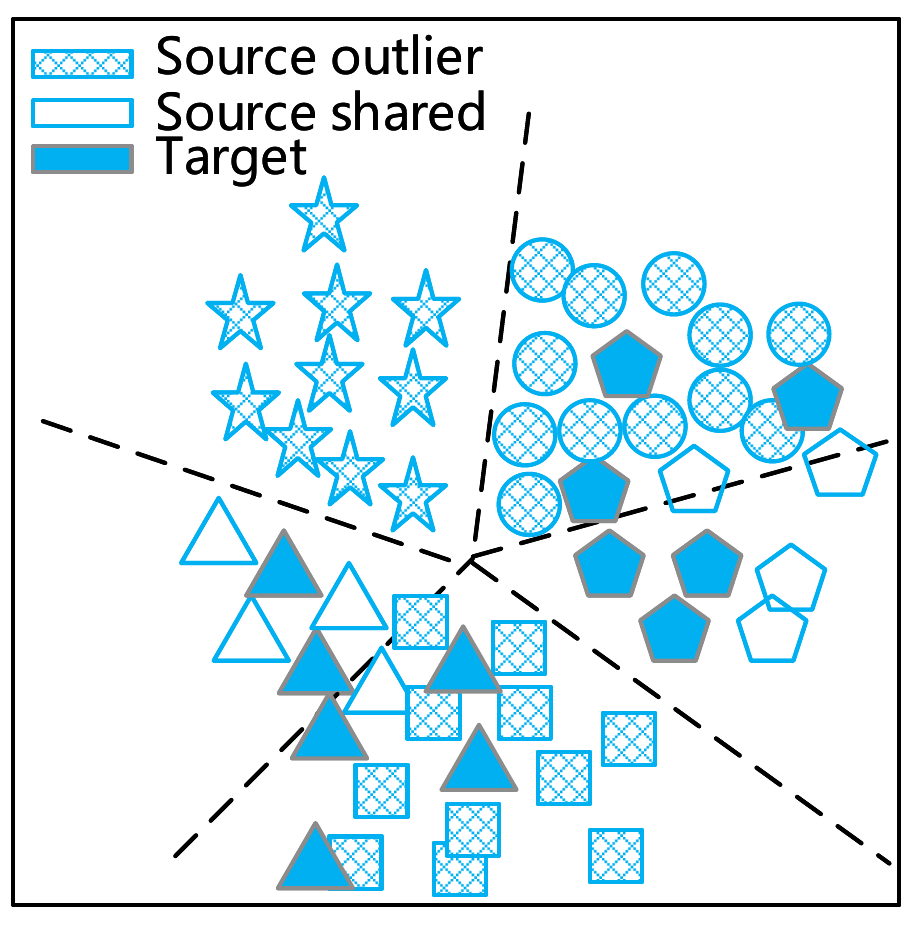}}
\end{minipage}}
\subfigure[]{\label{fig:diagram1_b}
\begin{minipage}[c]{0.33\textwidth}
\centering \scalebox{0.5}{
\includegraphics{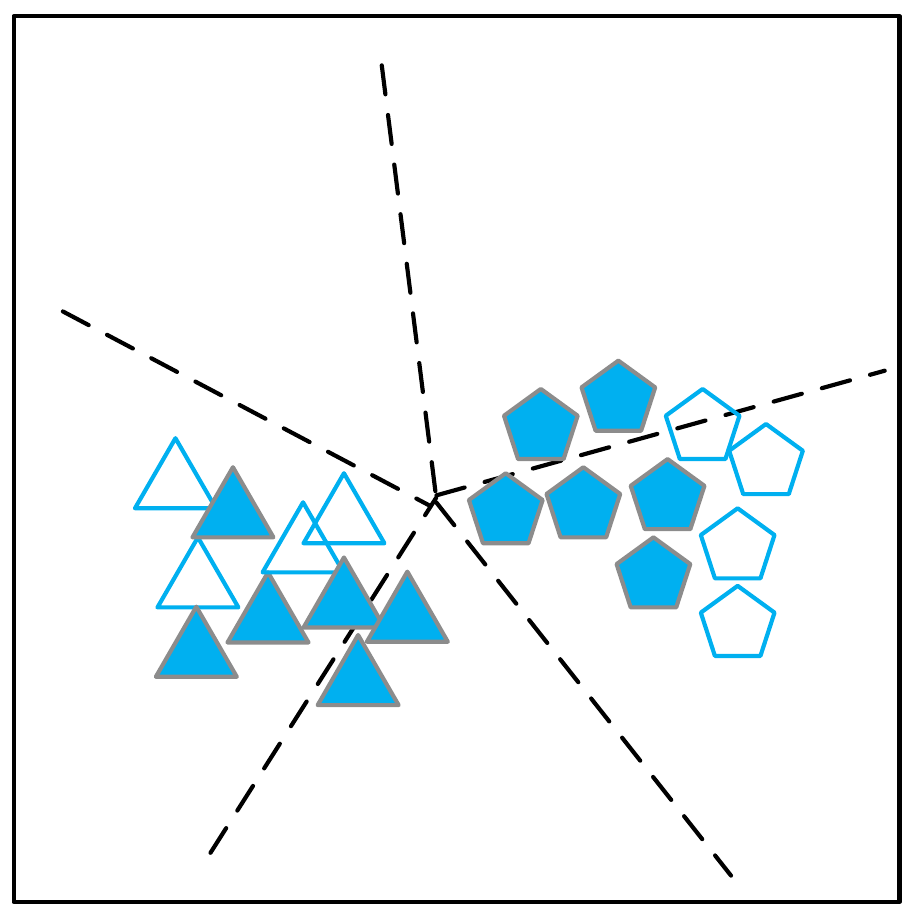}}
\end{minipage}}
\subfigure[]{\label{fig:diagram1_c}
\begin{minipage}[c]{0.33\textwidth}
\centering \scalebox{0.5}{
\includegraphics{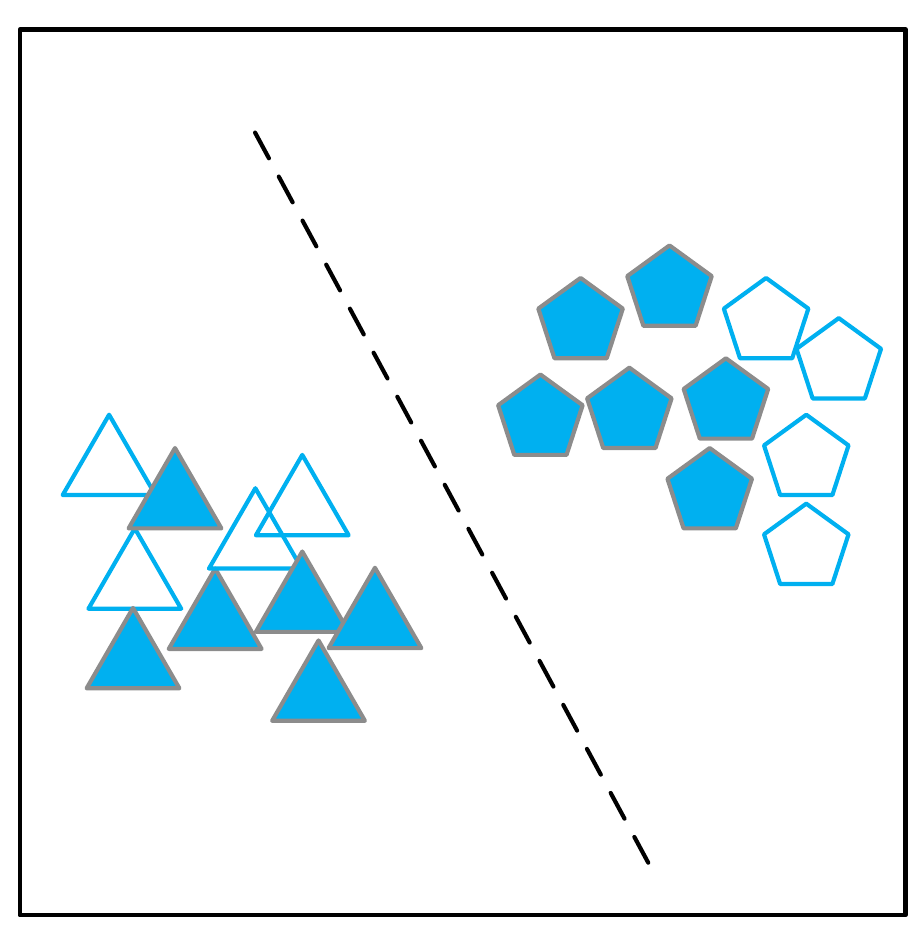}}
\end{minipage}}
\caption{Illustrations of negative transfer and classifier shift. (a) In a PDA task, UDA methods are likely to map some target samples into the feature space of source outliers, leading to negative transfer. (b) Existing PDA methods may suffer from label space shift. The classifier pays more attention to circle and square outliers due to their large sample size, and as a result, classifies triangles and pentagons (in the target domain) incorrectly. (c) TSCDA exploits a target-specific classifier to deal with label space shift, which effectively improves classification accuracy in the target domain.}\label{fig:diagram1}
\end{figure*}

Many UDA algorithms learn a domain-invariant and transferable feature space by matching the feature spaces of the source and target domains~\cite{DAN,zhang2019manifold,JAN_MMD,DANN,ADDA,RTN}. Their classification performance would drop greatly in the PDA scenario, as the intrinsic conditional distribution changes dramatically from one to the other. These models may incorrectly match the target samples to the source outliers, which is called \textit{negative transfer} in prior literature~\cite{PADA,IWAN,TWINS}. As depicted in Fig.~\ref{fig:diagram1_a}, some target samples (triangles and pentagons) are misclassified to the source-outlier classes (squares and circles). An intuitive strategy to address this problem is reweighing the samples in the source domain, through endowing source-shared samples with large weights while source-outliers with small weights during transferring. Then, domain-invariant features are learned from the reweighed source domain and the target domain, and a shared classifier is also learned to complete the classification knowledge transfer.

Following this inspiration, several methods deal with PDA tasks by weighed cross-entropy loss~\cite{TWINS} or weighed adversarial training~\cite{PADA,IWAN}, with good performance achieved. However, the weighed cross-entropy loss based methods ignore the alignment of feature spaces, which is considered the key to success of domain adaptation~\cite{domain_theory}, and the weighed adversarial training based methods may suffer training instability and mode collapse. Besides, these methods ignore the classifier incompatibility problem caused by the huge label space shift between different domains in PDA scenario. Consequently, the optimal classifier of the source domain is sub-optimal for the target domain. This problem is also called \textit{classifier shift}, as the classification over the source domain is a $|\mathcal{Y}^s|$-class problem while that over the target domain is a $|\mathcal{Y}^t|$-class problem. As shown in Fig.~\ref{fig:diagram1_b}, the target domain has only two categories while the source classifier is trained to classify five classes, and some samples of the target domain may be mis-classified by the source classifier. A more suitable solution is to classify the target domain samples by the target specific classifier as Fig.~\ref{fig:diagram1_c}. Therefore, the strategy of sharing a classifier between the two domains may be invalid in PDA problems. Furthermore, there are some target samples locating outside the support of \textit{source-shared-domain}, and they also contain discriminant information that benefits classification in the target domain. However, such information is often ignored.

In this paper, we propose a novel deep learning model for addressing PDA problem, termed Learning Target Domain Specific Classifier for Domain Adaptation and abbreviated as \textbf{TSCDA}. It contains two interdependent modules, a partial features alignment module and a target-specific classifier learning module. In the first module, a Soft Weighed Maximum Mean Discrepancy (\textbf{SWMMD}) is proposed to mitigate the negative transfer between the \textit{source-outlier-domain} and the target domain. The feature distributions of the \textit{source-shared-domain} and the target domain are therefore aligned in the feature space. The weights are automatically calculated using the source classifier on the target data, which indicate the probabilities of a source class appears in the \textit{source-shared-domain}. In the second module, TSCDA learns a target-specific classifier based on the pseudo-labels of high-confidence samples in the target domain. In particular, we propose a PEers Assisted Learning (\textbf{PEAL}) method to improve the classification performance and alleviate the over-fitting problem of the target classifier. PEAL minimizes the difference between prediction outputs of the target domain-specific and auxiliary classifiers, i.e., consistency loss. We conduct extensive experiments on three PDA benchmarks to evaluate effectiveness of the proposed TSCDA, and the results show it achieves state-of-the-art performance. Our main contributions are summarized as follows.
\begin{itemize}
	\item[1)] A novel deep learning method is proposed for tackling unsupervised PDA. It includes partial feature alignment and target-specific classifier learning, and can simultaneously alleviate negative transfer and classifier shift between the source and target domains.
	\item[2)] A soft weighed maximum mean discrepancy loss, SWMMD, is proposed to do partial feature matching. It essentially alleviates the possibility of negative transfer between the source-outlier-domain and the target domain.	
	\item[3)] A novel target-specific classifier is used to address the classifier shift problem. It aims to learn a more powerful classifier for the target domain. Moreover, a peers assisted learning approach, PEAL, is proposed to learn the target-specific classifier and improve the discrimination of the model.
\end{itemize}

The rest of this paper is organized as follows. In Section~\ref{sec:related work}, related works are briefly reviewed. In Section~\ref{sec:model}, the TSCDA algorithm is described in detail. Section~\ref{sec:experiments} presents extensive experiment results and validates the effectiveness of TSCDA. Section~\ref{sec:conclusion} concludes this paper.

\section{Related works}\label{sect:related-works}\label{sec:related work}

\subsection{Unsupervised Domain Adaptation}\label{subsec:relatedwork_UDA}

UDA aims to transfer knowledge from a labeled source domain to an unlabeled target domain. Recent study~\cite{domain_theory} has shown reducing the distribution discrepancy of the two domains can help learn domain-invariants with good discrimination on the target domain. Existing UDA methods can generally be divided to statistical feature alignment based methods~\cite{MMD,JAN_MMD,RTN,DAN,WMMD,CMD,CORAL} and adversarial learning based methods~\cite{ADDA,pixel-levelDA,DAN,GANbased1}.

The statistical feature alignment based methods reduce distribution discrepancy by minimizing the difference of statistical moments between the source and target domains~\cite{ren2016CGMMD,KLN}. Maximum mean discrepancy~(MMD)~\cite{Gretton2012MMD} is the most commonly used method for statistical feature matching. It approximately characterizes the difference between the two domains by using their respective empirical expectation. Due to its flexible modeling and simple optimization, several extensions~\cite{MMD,WMMD,JAN_MMD} have been developed to improve UDA performance and applied to other problems. Yan \etal~\cite{WMMD} presented a weighed MMD (WMMD) to align feature distributions with an unweighed loss, where the weight is calculated by the ratio of prior distributions for the two domains. Tzeng \etal~\cite{MMD} proposed a confusion loss based on MMD to align the domains. Long \etal~\cite{JAN_MMD} proposed a joint MMD principle to directly match the joint distributions. In addition, Zellinger \etal~\cite{CMD} minimized the domain discrepancy by matching the higher order central moments of domain-specific probability distributions. Sun \etal~\cite{CORAL} proposed to minimize domain bias by matching the mean and covariance of the distributions simultaneously. Some other works also explore weighted strategies in domain alignment~\cite{Ding-graph,Ding18}. Ding \etal~\cite{Ding-graph} proposed an iterative refinement scheme to optimize the probabilistic and class-wise subspace adaptation term from a graph-based label propagation perspective. In~\cite{Ding18}, two coupled deep neural networks were used to extract the representative features, then a weighted class-wise adaptation scheme was built for minimizing the distribution gap. 

With the development of Generative Adversarial Nets~\cite{GAN,WGAN}, adversarial learning based methods~\cite{ADDA,pixel-levelDA,GTA2018,DAN,GANbased1} achieve significant advances in UDA. These methods train a domain classifier to distinguish whether a sample is from source or target, and train a feature extractor to minimize distribution divergence between the two domains. Tzeng \etal~\cite{ADDA} developed a unified framework for UDA based on adversarial learning. Bousmalis \etal~\cite{pixel-levelDA} proposed to learn the transformation between different domains on pixel level. Saito \etal~\cite{MCD} proposed two classifiers to consider task-specific decision boundaries by maximizing the difference between the outputs of the two classifiers.

In this work, we present a SWMMD module, which is a new weighed MMD, to mitigate negative transfer between the \textit{source-outlier-domain} and the target domain, and align feature distributions of the \textit{source-shared-domain} and the target domain. Note that SWMMD differs from WMMD~\cite{WMMD} in two aspects. First, SWMMD uses the probabilistic outputs to estimate the weights, while WMMD uses the hard labels, which would increase uncertainty in both pseudo-labeling and domain alignment. Second, the weights of SWMMD are designed for differentiating the \textit{source-shared-domain} from the \textit{source-outlier-domain}, while WMMD only applies to balance the class-wise distributions between source and target domains. Thus, the class prior probabilities of the source domain are not incorporated into the weights of SWMMD. 

\begin{figure*}[t]
	\centering
	\includegraphics[width=0.85\textwidth]{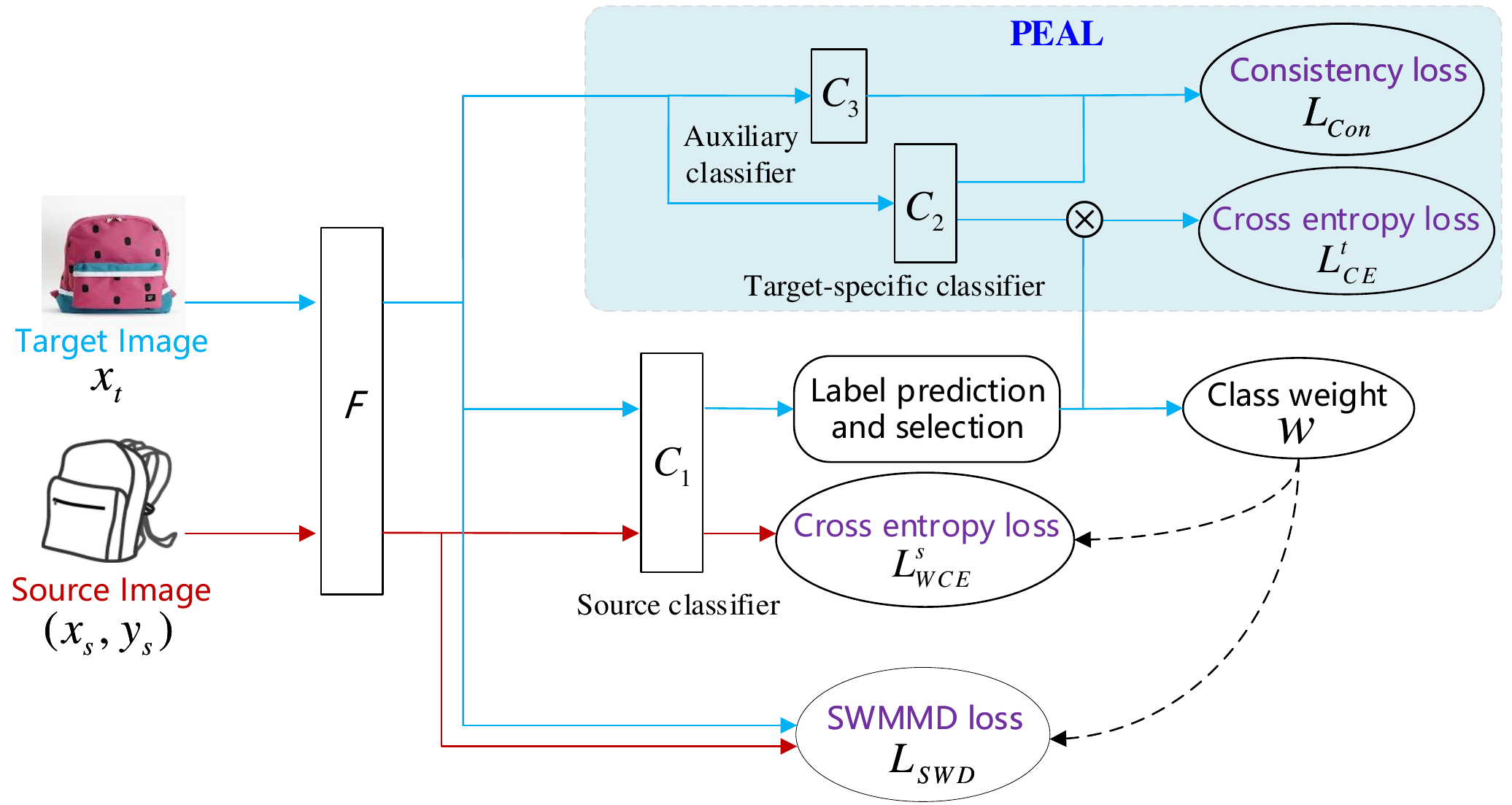}
	\caption{Workflow of TSCDA. The blue line and the red line represent the forward propagation of target domain and source domain, respectively. The source classifier $C_1$ is trained with a weighed cross-entropy loss. Considering label space shift between the two domains, the target classifier $C_2$ is trained with the target data and their pseudo-labels predicted by $C_1$. The PEAL module is trained with a consistency regularization term, which uses the outputs of $C_2$ and $C_3$. At last, the feature extractor $F$ is trained by minimizing all loss functions. Best viewed in color.}
	\label{fig:flowchart}
\end{figure*}

\subsection{Partial Domain Adaptation}\label{subsec:relatedwork_PDA}

In PDA scenario, many UDA methods suffer noticeable performance degradation due to the notorious negative transfer, especially when the target label space is much smaller than the source label space. Most UDA works build a mechanism to reweigh source samples so that the reweighed source label distribution approximates the target label distribution. In other words, they attempt to reformulate the PDA problem to a UDA problem. Cao \etal~\cite{PADA} proposed to decrease the contributions of source-outlier classes, and increase those of source-shared classes by weighting source samples under an adversarial framework~\cite{DANN}. It effectively promotes positive transfer and alleviates negative transfer between the two domains. Zhang \etal~\cite{IWAN} presented a two-domain-classifier strategy to identify the importance scores of source samples, and match the reweighed source domain and the target domain based on adversarial learning. In addition, Xu \etal~\cite{AFN} proposed to adapt feature norms of both domains to achieve equilibrium, which is free from relationship of the label spaces. Recently, some methods apply ensemble learning to improve the discrimination ability of extracted features~\cite{SelfEnsembling,CAT}. French \etal~\cite{SelfEnsembling} used a mean teacher model~\cite{meanteacher} to mine the target domain knowledge. Deng \etal~\cite{CAT} explored the class-conditional structure of the target domain with an ensemble teacher model.

In this work, besides the feature alignment module, we also propose an unshared classifier method to deal with classifier shift. Specifically, we adopt a discussion strategy to design the classifier learning module (i.e., PEAL), which minimizes the output inconsistency between the target-specific classifier and auxiliary classifiers. We can initialize several (at least two) classifiers in the target domain and then transfer the decision knowledge by alternatively optimizing them until all of them are competitive. Note that these classifiers are parallel, rather than contrastive as those in the Teacher-Student model~\cite{hinton2015distilling}.

The utilization of the inconsistency loss connects our PEAL to other UDA methods~\cite{Ding-graph,MCD,TWINS}, but their motivations and usages are different. In~\cite{Ding-graph}, inconsistency loss is used for subspace adaptation. Saito~\etal~\cite{MCD} maximized the discrepancy of the task-specific classifiers for effectively measuring the domain distance. However, PEAL minimizes the classifier inconsistency to deal with the label space shift problem in PDA. Matsuura~\etal~\cite{TWINS} also minimized the classifiers¡¯ inconsistency on target samples. Note that the classifiers are trained on the source domain, instead of target-specific. In PEAL, the inconsistency loss is used to facilitate the collaborative learning between multiple classifiers for improving the generalization ability of the target-specific classifier.

\section{Our Method}\label{sec:model}

\subsection{Notations and Overall Structure}\label{subsec:motivation}

Several important notations are defined here. The source data $\mathcal{X}^s\in \mathbb{R}^{d\times n_s}$ and the target data $\mathcal{X}^t\in \mathbb{R}^{d\times n_t}$ are drawn respectively from the source domain distribution $p_s(\mathbf{x})$ and the target domain distribution $p_t(\mathbf{x})$, where $d$ is the data dimension, and $n_s$, $n_t$ represent the sample sizes of source and target domains. The label vector of source domain is formulated as $\mathcal{Y}^s$. Due to domain shift, the distributions of source and target domains are different but similar, i.e. $p_s(\mathbf{x})\neq p_t(\mathbf{x})$. Besides, $p_i(y|\mathbf{x})$ denotes the predicted category distribution of $\mathbf{x}$ by classifier $C_i$, $i\in \{1,2,\cdots,n\}$. Specifically, $C_1$ is the source classifier, $C_2$ is the target-specific classifier, and $C_3,\cdots,C_n$ are auxiliary classifiers. We use $F$ to denote the shared deep feature extractor.

The proposed TSCDA method contains two interdependent modules, i.e., partial features alignment and target-specific classifier learning. The overview of TSCDA is shown in Fig.~\ref{fig:flowchart}. On the one hand, in order to control the negative transfer between the \textit{source-outlier-domain} and the target domain, partial features alignment module uses SWMMD to partially align the feature distributions of the \textit{source-shared-domain} and target domains. On the other hand, to consider the label space shift between the source and target domains, the target-specific classifier learning module learns a target-specific classifier $C_2$ for the target domain by the pseudo-labels. Specially, PEAL is implemented by a consistency regularization between the outputs of the target-specific classifier $C_2$ and the auxiliary classifier $C_3$.


\subsection{Partial Features Alignment}\label{subsec:partial feature alignment}

Since the existence of the \textit{source-outlier-domain} seriously mislead knowledge transfer between the source and target domains, the first goal of our method is to identify and distinguish the \textit{source-shared-domain} and the \textit{source-outlier-domain}, then filter out samples in the \textit{source-outlier-domain}. It is achieved by a soft weight vector calculated by probabilistic outputs of target samples on the source classifier $C_1$,
\begin{equation}\label{eqn:sample weight}
\mathbf{w} = \frac{1}{n_t}\sum_{\mathbf{x}_j\in \mathcal{D}^t}p_1(y|\mathbf{x}_j),
\end{equation}
where $\mathbf{w}\in \mathbb{R}^{|\mathcal{Y}^s|}$ is initialized with $[1/|\mathcal{Y}^s|,\dots,1/|\mathcal{Y}^s|]$. A large $\mathbf{w}_i$ value indicates a high probability that the $i$-th class of the source domain appears in the target domain. Thus, $\mathbf{w}_i$ can also be viewed as the probability that the $i$-th class belongs to the \textit{source-shared domain}.

Although MMD has been widely used to solve UDA tasks by matching feature distributions of the two domains, it aligns the whole distribution of the source domain with that of the target domain, and the source-outliers are harmful to the correct knowledge transfer in PDA tasks. To mitigate negative transfer, SWMMD is proposed by weighting MMD with the soft weight vector $\mathbf{w}$. Let $F(\cdot)$ be the mapping function of the feature extractor and $\Phi(\cdot)$ a nonlinear kernel mapping. We denote $z_w\!=\!\sum_{x_{i} \in \mathcal{D}^{s}} \mathbf{w}_{y_{i}}$ for simplicity. The objective function can be formulated as
\begin{equation*}
  L_{SW\!D}(\mathbf{X}^s,\mathbf{X}^t)\!=\!\|\frac{1}{n_t}\!\!\sum_{\mathbf{x}_j\in \mathcal{D}^t}\!\!\Phi\circ F(\mathbf{x}_j)-\frac{1}{z_w}\!\!\sum_{\mathbf{x}_i\in\mathcal{D}^s}\!\!\mathbf{w}_{y_i}\Phi\circ F(\mathbf{x}_i)\|_{\mathcal{H}}^2.
\end{equation*}

We define kernel function $k$ as $k(\mathbf{x}_i,\mathbf{x}_j)=\langle\Phi(\mathbf{x}_i),\Phi(\mathbf{x}_j)\rangle$. Then the loss function can be rewritten as
\begin{eqnarray}
\label{eqn:WMMD_expand}
\begin{aligned}
& L_{SW\!D}(\mathbf{X}^s,\mathbf{X}^t) = \frac{1}{n_t^2}\sum_{\mathbf{x}_j\in \mathcal{D}^t}\sum_{\mathbf{x}_j^{'}\in \mathcal{D}^t}k(F(\mathbf{x}_j),F(\mathbf{x}_j^{'}))\\
& -\frac{2}{n_tz_w}\sum_{\mathbf{x}_j\in \mathcal{D}^t}\sum_{\mathbf{x}_i\in \mathcal{D}^s}\mathbf{w}_{y_i}k(F(\mathbf{x}_j),F(\mathbf{x}_i))\\
&+\frac{1}{z_w^2}\sum_{\mathbf{x}_i\in\mathcal{D}^s}\sum_{\mathbf{x}_i^{'}\in\mathcal{D}^s}\mathbf{w}_{y_i}\!\mathbf{w}_{y_i^{'}} k(F(\mathbf{x}_i),F(\mathbf{x}_i^{'})).
\end{aligned}
\end{eqnarray}

In this way, SWMMD aligns partially the feature distributions of the \textit{source-shared-domain} and the target domain to address domain shift, thus effectively boosts the positive transfer of the \textit{source-shared-domain} and reduces the negative transfer of the \textit{source-outlier-domain}.

\subsection{Learning Domain-specific Classifiers}\label{subsec: adaptive classifiers}

To reflect the importance of each category in training the source classifier, we use $\mathbf{w}$ learned from Formula~\eqref{eqn:sample weight} to reweigh the cross-entropy loss. In the weighed and supervised learning phase, classes with high probabilities are given large weights while classes absent in the target domain are given tiny weights. Specifically, the weighed cross-entropy loss is formulated as
\begin{equation}\label{eqn:weighed source supervised loss}
L_{WCE}^s(\mathbf{X}^s,Y^s)= -\frac{1}{n_s}\!\sum_{\mathcal{D}^s}\mathbf{w}_{y_i}\!\!\sum_{k=1}^{|\mathcal{Y}^s|}\mathbbm{1}_{[k=y_i]}\log p_1(k|\mathbf{x}_i),
\end{equation}
where $\mathbf{w}_{y_i}$ is the class-wise weight of source sample $\mathbf{x}_i$. Under this setting, the source classifier pays more attention to the classes in the source-shared-domain and thus can partially address label space shift. We denote the classifier induced by $L^s_{WCE}$ as $C_1$ hereinafter for simplicity.

Due to the classifier shift problem, the source classifier $C_1$ may be sub-optimal in the target domain, and thus we propose to learn a target-specific classifier. Following~\cite{CAT,PFAN}, we use $C_1$ to label target samples, as it performs well on a majority of target samples. To obtain reliable pseudo-labels, we impose a constraint that the confidence of the target sample predicted by $C_1$ should exceed a threshold $\nu$. It means that a target sample $\mathbf{x}_i^t$ will be tagged with a pseudo-label $\hat{y}_i^t$ if
 \begin{equation*}
   \max_{y}\: p_1(y|\mathbf{x}_i^t)\geq \nu.
 \end{equation*}
According to these pseudo labels, the target domain can be divided into two disjoint sub-domains, $\mathcal{D}^t=\mathcal{D}^{tl}\cup\mathcal{D}^{tu}$, where  $\mathcal{D}^{tl}=\{(\mathbf{x}_i^{tl}, \hat{y}_i^{tl})\}_{i=1}^{n_{tl}}$, $\mathcal{D}^{tu} = \{\mathbf{x}_i^{tu}\}_{i=1}^{n_{tu}}$ are target sets with pseudo-labels and without pseudo-labels, and $n_{tl}$, $n_{tu}$ are the number of instances in $\mathcal{D}^{tl}$ and $\mathcal{D}^{tu}$, respectively.

With the pseudo-labeled target set $\mathcal{D}^{tl}$, the cross-entropy loss is used again to train the target-specific classifier $C_2$, i.e.,
\begin{eqnarray}
\label{eqn:target classifer loss}
\begin{aligned}
L_{CE}^{t}(\mathbf{X}^{tl},\hat{Y}^{tl})=-\frac{1}{n_{tl}}\sum_{\mathbf{x}_i\in \mathcal{D}^{tl}}\sum_{k=1}^{|\mathcal{Y}^s|}\mathbbm{1}_{[k=\hat{y}_i]}\log p_2(k|\mathbf{x}_i).
\end{aligned}
\end{eqnarray}
We denote the classifier induced by $L^t_{CE}$ as $C_2$ hereinafter.

By combining Formula~\eqref{eqn:weighed source supervised loss} and~\eqref{eqn:target classifer loss}, the classification loss $L_{CE}$ of TSCDA can be written as
\begin{eqnarray}
\label{eqn:whole classification loss}
\begin{aligned}
L_{CE}=L_{WCE}^s+L_{CE}^t.
\end{aligned}
\end{eqnarray}

\subsection{Peers Assisted Learning}

\begin{figure}
\centering
\includegraphics[width=0.9\columnwidth]{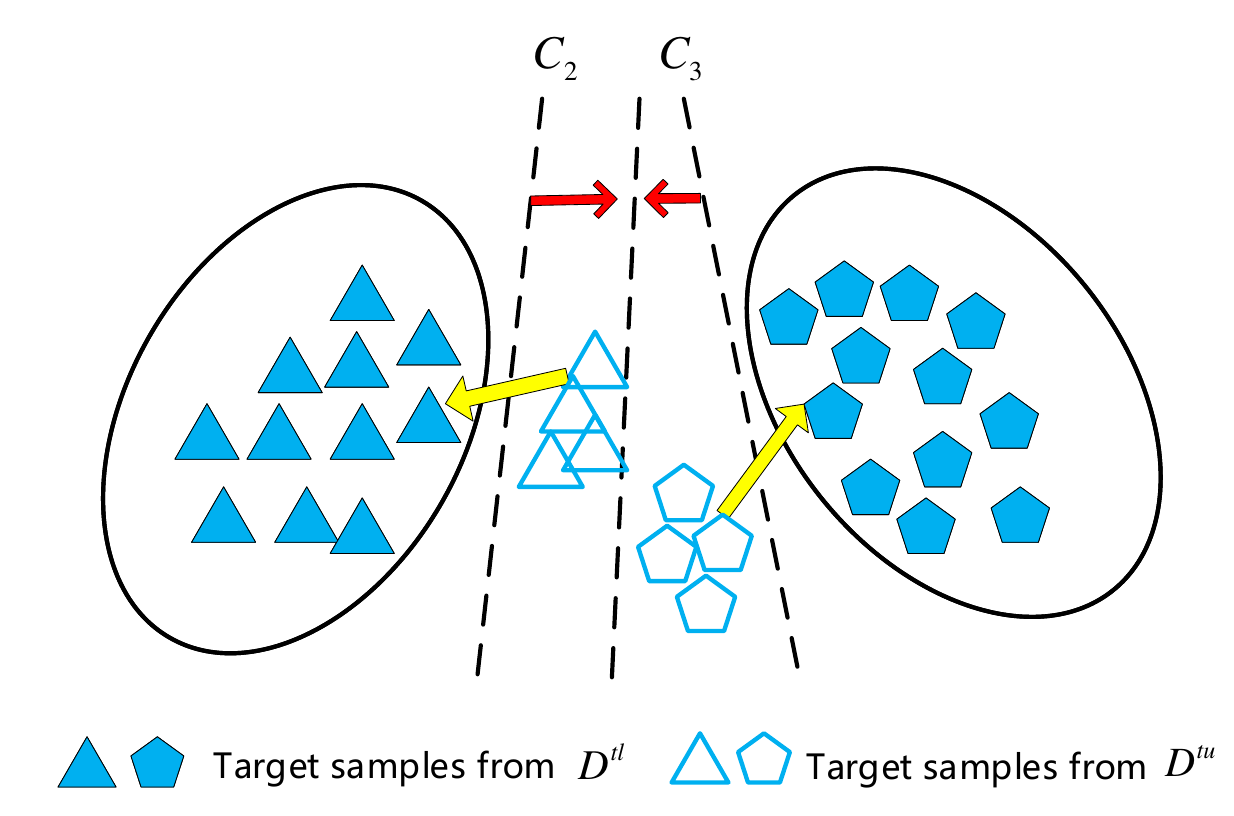}
\caption{The working principle of PEAL. On one hand, the inconsistency loss impels target-specific classifier $C_2$ and auxiliary classifier $C_3$ to learn from each other, forcing their decision boundaries to move toward each other. On the other hand, the inconsistency loss pushes samples lying in the classification contradiction zone close to their class centers.}
\label{fig:consistency}
\end{figure}

Since the target classifier $C_2$ is only trained on $\mathcal{D}^{tl}$, target samples from $\mathcal{D}^{tu}$ are likely to be incorrectly classified when the classifier is over-fitting on $\mathcal{D}^{tl}$, as shown in Fig.~\ref{fig:consistency}. To improve the generalization ability of $C_2$ and the adaptability of the decision boundary to the target domain, we present an auxiliary classifier $C_3$, with a consistency regularization term. The inconsistency loss is used to minimize the mean difference between the predicted category distributions, which are outputs of classifiers $C_2$ and $C_3$.

Intuitively, the learning process between $C_2$ and $C_3$ can be regarded as the discussion and exchange between peers, thus they play similar roles in reducing the inconsistent loss. It is desirable that classifiers $C_2$ and $C_3$ can learn from each other, and we call this module Peers Assisted Learning~(PEAL). Several (conditional) distribution measurements, such as Kullback-Leibler divergence and Wasserstein distance~\cite{WGAN}, can be exploited here to measure the difference or inconsistency. We use the Euclidean distance in this work due to its simple numerical optimization. The inconsistency loss between classifiers $C_2$ and $C_3$ can be written as
\begin{eqnarray}
\label{eqn:consistency loss}
\begin{aligned}
L_{Con}(\mathbf{X}^t)= \frac{1}{n_t}\sum_{\mathbf{x}_j\in \mathcal{D}^t}\|p_2(y|\mathbf{x}_j)-p_3(y|\mathbf{x}_j)\|^2_2.
\end{aligned}
\end{eqnarray}

The working principle of PEAL is shown in Fig~\ref{fig:consistency}. The consistency regularization term forces the target-specific classifier $C_2$ and the auxiliary classifier $C_3$ to learn from each other, in this way their decision boundaries move toward consensus and finally reach an equilibrium. At this time, $C_2$ and $C_3$ are competitive and assimilative to each other. However, compared with its initial state, the generalization ability of $C_2$ is improved. In addition, the inconsistency loss promotes target samples locating at the contradiction area of the decision boundaries to move toward the support of their classes in the target domain, which further improves the discrimination power of the classifiers.

The inconsistency loss can be extended to multiple auxiliary classifiers $C_3,\cdots,C_n$, i.e.,
\begin{equation}\label{eqn:multiple consistency}
\tilde{L}_{Con}(\mathbf{X}^t) = \frac{1}{n_t}\sum_{\mathbf{x}_j\in\mathcal{D}^t}\sum_{m_1\neq m_2} \|p_{m_1}(y|\mathbf{x}_j)-p_{m_2}(y|\mathbf{x}_j)\|^2_2.
\end{equation}

Note that even though the PEAL module exploits multiple auxiliary classifiers to improve discriminant performance of the target-specific classifier, it is essentially different from the ensemble learning methods~\cite{SelfEnsembling,kim2018attention}. PEAL uses mutual learning to achieve collaborative improvement, while ensemble methods focus on classifier reinforcement via dynamically updating weights of both training samples and sub-classifiers.

With an alternative classifier learning strategy, our method is also similar to the Teacher-Student model~\cite{meanteacher} for knowledge distillation and model compression. Generally, the latter is used to transfer knowledge from a complex (Teacher) network to a simple (Student) network. Only the student network is trained in an iterative manner to learn useful information from the Teacher network as much as possible. Empirical results of Hinton et al.~\cite{hinton2015distilling} have shown that the knowledge distillation scheme can offer student networks with comparable accuracy of the Teacher model. However, both motivation and mechanism of PEAL are different from the Teacher-Student model. The main difference is that the unequal status between teacher and student is eliminated from PEAL.

\begin{algorithm}
	\caption{The TSCDA Algorithm.}
	\label{alg:TSCDA}
	\begin{algorithmic}[1]
		\REQUIRE  Data: source domain $\mathcal{D}^s=\{(\mathbf{x}_i^s,y_i^s)\}_{i=1}^{n_s}$ and target domain $\mathcal{D}^t=\{\mathbf{x}_j^t\}_{j=1}^{n_t}$. Networks: feature extractor $F$, source classifier $C_1$, target classifier $C_2$ and auxiliary classifier $C_3$. Parameters: confidence threshold $\nu$, trade-off coefficients $\beta$, $\gamma$.	
		\ENSURE Updated $F$ and $C_2$.		
		\STATE \textbf{Initialize:} $\mathbf{w}=[1/|\mathcal{Y}^s|,\dots,1/|\mathcal{Y}^s|]\in \mathbb{R}^{|\mathcal{Y}^s|}$.
		\STATE Pre-train: optimize $F$ and $C_1$ by Eq.~\eqref{eqn:weighed source supervised loss};		
        \WHILE{not converged}
		\FOR {\textbf{each batch}}
		\STATE Minimize the loss Eq.~\eqref{eqn:weighed source supervised loss} w.r.t. $F$ and $C_1$;
        \STATE Minimize the loss Eq.~\eqref{eqn:WMMD_expand} w.r.t. $F$;
        \STATE Label the target data by $F$ and $C_1$;
        \STATE Minimize the loss Eq.~\eqref{eqn:target classifer loss} and loss Eq.~\eqref{eqn:consistency loss} w.r.t. $F$, $C_2$ and $C_3$;
        \ENDFOR
		\STATE Update sample weight $\mathbf{w}$ by $F$ and $C_1$ with Eq.~\eqref{eqn:sample weight}.
		\ENDWHILE
	\end{algorithmic}
\end{algorithm}

\subsection{Model Training and Test}\label{subsec:algorithm}

Based on the above discussion, the classification loss $L_{CE}$ enables TSCDA to learn a source classifier and a target-specific classifier with discriminant feature representation. The inconsistency loss $L_{Con} $ promotes the target-specific classifier to alleviate over-fitting problem on the pseudo-labeled target domain. The partial feature alignment loss $L_{SW\!D}$ controls the negative transfer between the source-outlier-domain and the target domain. Combining Formula~\eqref{eqn:WMMD_expand},~\eqref{eqn:whole classification loss} and~\eqref{eqn:consistency loss}, the final objective function of TSCDA is
\begin{eqnarray}
\label{eqn:Total loss}
\mathcal{L}=L_{CE} + \beta L_{Con} +\gamma L_{SW\!D},
\end{eqnarray}
where $\beta$ and $\gamma$ are positive trade-off coefficients.

The most important component of TSCDA is the target classifier, with performance relying on the pseudo-labeling accuracy of the source classifier. We pre-train the feature extractor $F$ and the source classifier $C_1$ on the source domain, with cross entropy loss $L_{WCE}^s$, to ensure the accuracy and reliability of pseudo-labeling. Then the total loss (\ref{eqn:Total loss}) is optimized by back propagation. At last, we update the source sample weight $\mathbf{w}$. These steps are repeated until pre-fined convergence conditions are achieved, so as to obtain the feature extractor $F$ and the target-specific classifier $C_2$. The whole training algorithm is presented in Algorithm~\ref{alg:TSCDA}.

In the test stage, given a target sample $\mathbf{x}^t$, we first extract the feature of $\mathbf{x}^t$ by $F$. Then the target-specific classifier $C_2$ is used to predict the labels via
\begin{equation*}
  \hat{y}^t = \argmax C_2(F(\mathbf{x}^t)).
\end{equation*}
In particular, there are three classifiers, $C_1$, $C_2$, and $C_3$, and each one can be used to classify samples in the target domain. In Section~\ref{sec:experiments}, we will conduct experiments to evaluate the classification performance of different classifiers.

\section{Experiment Results and Analysis}\label{sec:experiments}

We conduct extensive experiments on three benchmark datasets to evaluate the effectiveness of our TSCDA through comparisons with some state-of-the-art UDA and PDA methods. All experiments are performed in the unsupervised PDA settings. Besides, we also perform ablation studies to show the effects of key components of the proposed method.

\subsection{Experiment Setup and Implementation Details}\label{subsec:setup}

\subsubsection{Datasets}

\begin{figure*}[htb]
\centering
\includegraphics[width=0.9\textwidth]{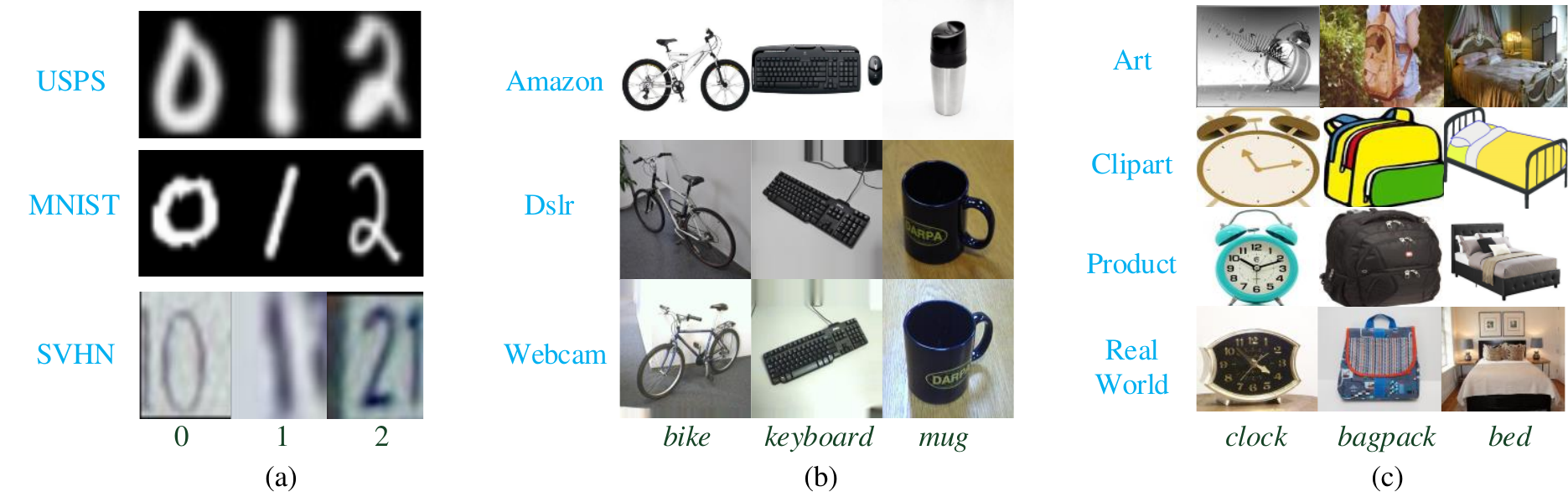}
\caption{Example images used in the experiments. (a) Digits. (b) Office-31. (c) Office-Home.}\label{fig:Datasets}
\end{figure*}

The datasets are briefly introduced here.
\begin{itemize}
	\item \textbf{\textit{Digits }} We build PDA experiments with the full training sets of MNIST~\cite{LeNet}, USPS~\cite{USPS} and SVHN~\cite{SVHN} as three different domains. Some exemplar images are shown in Fig.~\ref{fig:Datasets}(a). Each domain consist of ten classes of digits. In each transfer task, the source domain contains all images while the target domain contains only the first five classes following an ascending order of the source domain~\cite{TWINS}. We conduct three PDA tasks (\textbf{MNIST}$\rightarrow$\textbf{USPS}, \textbf{USPS}$\rightarrow$\textbf{MNIST}, and \textbf{SVHN}$\rightarrow$\textbf{MNIST}) on this dataset.

 \item \textbf{\textit{Office-31 }}\cite{office31} It includes a total of $4,652$ images of $31$ categories from three domains: \textit{Amazon} (\textbf{A}), \textit{DSLR} (\textbf{D}) and \textit{Webcam} (\textbf{W}), which are website images, digital SLR camera images and web camera images, respectively. Examples from this dataset are shown in Fig.~\ref{fig:Datasets}(b). Following the setting of~\cite{PADA}, we take samples of ten classes shared by \textit{Office-31} and \textit{Caltech-256}~\cite{caltech} as target domains. We evaluate the methods across all the six transfer tasks, i.e., \textbf{A}$\rightarrow$\textbf{D}, \textbf{A}$\rightarrow$\textbf{W}, $\cdots$, \textbf{W}$\rightarrow$\textbf{D}.

\item \textbf{\textit{Office-Home }}\cite{officehome} It is a more challenging dataset for visual domain adaptation. It contains four domains: \textit{Artistic} (\textbf{Ar}), \textit{Clipart} (\textbf{Cl}), \textit{Product} (\textbf{Pr}) and \textit{Real-World} (\textbf{Rw}), which are artistic images, real-world images, clipart images and product images, respectively. They have a total of $15,500$ images in $65$ categories of daily objects. Some examples are shown in Fig.~\ref{fig:Datasets}(c). To be consistent with protocols in~\cite{PADA,AFN}, we use the first $25$ categories following the alphabetic order to form target domains. We conduct all the twelve transfer tasks, i.e., \textbf{Ar}$\rightarrow$\textbf{Cl}, \textbf{Ar}$\rightarrow$\textbf{Pr}, $\cdots$, \textbf{Rw}$\rightarrow$\textbf{Pr}.
\end{itemize}

\begin{table*}[htb]
	\caption{Accuracy(\%) on Digits dataset for Partial unsupervised domain adaptation.}
	\label{tab:Acc_Digits}
	\centering
    \renewcommand{\tabcolsep}{1.3pc} 
    \renewcommand{\arraystretch}{1.2} 
		\begin{tabular}{cccccc}
			\hline 
			\multicolumn{2}{c}{Method} &  MNIST$\rightarrow$USPS  &  USPS$\rightarrow$MNIST  & SVHN$\rightarrow$MNIST  & Avg   \\
			\hline
			\multirow{3}{0.3in}{UDA} & Source Only  & $85.2$   & $80.0$   &  $73.9$  &  $79.7$   \\
			& DAN~\cite{DAN}  & $83.5$   & $80.7$   &  $70.9$  &  $78.4$  \\
			& DANN~\cite{DANN}  & $67.1$   & $72.1$   &  $39.8$  &  $59.7$ \\
			\hline
			\multirow{2}{0.3in} {PDA}& IWAN~\cite{IWAN}  & $90.6$   & $85.7$   &  $75.6$  &  $84.0$  \\
			&PADA~\cite{PADA}  & $90.6$   & $86.7$   &  $81.4$  &  $86.2$  \\
			\hline
			\multirow{1}{0.3in}{Ours} & {\textbf{TSCDA}}  & $\mathbf{97.4}\pm0.2$   & $\mathbf{90.1}\pm0.6$   &  $\mathbf{84.5}\pm3.8$  &  $\mathbf{90.7}$ \\
			\hline 
	\end{tabular}
\end{table*}

\subsubsection{Baselines}

To evaluate the effectiveness of TSCDA, we compare its performance with the baseline method, which fine-tunes ResNet-50 on the source domain, and also several state-of-the-art UDA and PDA methods based on deep network architectures, including Deep Adaptation Network (DAN)~\cite{DAN}, Domain Adversarial Neural Network (DANN)~\cite{DANN}, Adversarial Discriminative Domain Adaptation (ADDA)~\cite{ADDA}, Residual Transfer Networks (RTN)~\cite{RTN}, Joint Adaptation Network (JAN)~\cite{JAN_MMD}, Selective Adaptation Networks (SAN)~\cite{SAN}, Partial Adversarial Domain Adaptation (PADA)~\cite{PADA}, Importance weighed Adversarial Nets (IWAN)~\cite{IWAN}, Two weighed Inconsistency-reduced Network (TWINs)~\cite{TWINS} and Adaptive Feature Norm (AFN)~\cite{AFN}. Among them, DAN, DANN, ADDA, RTN, JAN and AFN are UDA methods, while SAN, PADA, IWAN, TWINs and AFN are PDA methods.

We perform ablation study with three variants, i.e., \textbf{TSCDA-v1}, \textbf{TSCDA-v2} and \textbf{TSCDA-v3}, which are built by subtracting from TSCDA the target-specific classifier learning module, the PEAL module and the SWMMD module, respectively.

\subsubsection{Implementation Details}

Following the standard protocols~\cite{PADA,IWAN,AFN}, we use all labeled source data and all unlabeled target data to conduct experiments on PyTorch platform. The ResNet-50 model pre-trained on ImageNet~\cite{Imagenet} is adopted as the initial feature extractor $F$.

SWMMD is placed onto the last layer of $F$, and the weights of SWMMD are updated in each epoch. A Gaussian kernel with bandwidth $1$ is used to calculate SWMMD. Classifiers $C_1$, $C_2$ and $C_3$ share the same architecture which places two fully connected layers (2048-512-$\omega$) onto the feature extractor $F$, where $\omega$ is the number of source domain classes. The classifiers are initially equipped with a Gaussian distribution $\mathcal{N}(0,0.05)$, and the activation function is LeakyReLU with a slope of $0.2$.

The Adam optimizer with $\beta_1=0.9$ and $\beta_2=0.999$ is uniformly used in mini-batch based optimization. The learning rate is 2$e$-4. The hyper-parameters are uniformly set to $\beta=0.1$, $\gamma=0.4$, and $\nu=0.9$. For inputs, we use $224\times 224$ center crops of $256\times 256$ resized images on these datasets. As for the experiments on \textit{Digits}, the images are uniformly resized to $32\times 32$. Although tuning hyper-parameters can improve the performance of each subtask, we do not tune parameters for each transfer task because it will greatly increase the computational cost.

\subsection{Comparison with State-of-the-Art Methods}\label{subsec:results}

We compare the performance of TSCDA against state-of-the-art UDA, PDA methods. We report the mean accuracy and standard deviation (STD) of TSCDA for twenty repeated random trials. The accuracies of state-of-the-art methods are directly cited from the original papers.

Table~\ref{tab:Acc_Digits} lists the results of our TSCDA for three transfer tasks on \textit{Digits}. To evaluate the source only model, we only use the data from source domain to train a network with the same architecture as our method. We also compared our methods with a series of UDA and PDA methods as shown in Table~\ref{tab:Acc_Digits}. On one hand, we observe that UDA methods cannot outperform the Source Only model. This is potentially because UDA methods transfer the incorrect knowledge of the \textit{source-outlier-domain} to the target domain by aligning the whole source distribution and the target distribution. Besides, PDA methods outperform traditional UDA methods with a large margin. It indicates that controlling negative transfer is critical for successful PDA. On the other hand, the mean accuracy of TSCDA for all three subtasks is $90.7\%$, which is much higher than that of the state-of-the-art methods by a large margin $4.5\%$. Besides, TSCDA outperforms the existing methods for all subtasks. Specifically, the classification accuracies of TSCDA on these three tasks are $94.7\%$, $90.1\%$ and $84.5\%$, which outperform the the state-of-the-art method PADA by $6.8\%$, $3.4\%$ and $3.1\%$. All of these results validate the effectiveness of our method.

\begin{table*}[htb]
	\caption{Accuracy(\%) on \textit{Office-31} dataset for PDA tasks.}
	\label{tab:Acc_office31}
	\centering
    \renewcommand{\tabcolsep}{0.8pc} 
    \renewcommand{\arraystretch}{1.2} 
	\scalebox{1.0}{
		\begin{tabular}{ccccccccc}
			\hline
			\multicolumn{2}{c}{Method} &  A$\rightarrow$W  &  D$\rightarrow$W  & W$\rightarrow$D  & A$\rightarrow$D  &  D$\rightarrow$A & W$\rightarrow$A & Avg   \\
			\hline
			\multirow{6}{0.3in}{UDA} & ResNet~\cite{ResNet}  & $54.52$   & $94.57$   &  $94.27$  &  $65.61$  & $73.17$  &  $71.71$  &  $75.64$ \\
			& DAN~\cite{DAN}  & $46.44$   & $53.56$   &  $58.60$  &  $42.68$  & $65.66$  &  $65.34$   &  $55.38$ \\
			& DANN~\cite{DANN} & $41.35$ & $46.78$ & $38.85$ & $41.36$ & $41.34$ & $44.68$ & $42.39$ \\
			& ADDA~\cite{ADDA}  & $43.65$   & $46.48$   &  $40.12$  &  $43.66$  & $42.76$  &  $45.95$   &  $43.77$   \\
			& JAN~\cite{JAN_MMD}  & $43.39$   & $53.56$   &  $41.40$  &  $35.67$  & $51.04$   &  $51.57$  &  $46.11$  \\
			& RTN~\cite{RTN}  & $75.25$   & $97.12$   &  $98.32$  &  $66.88$  & $85.59$  &  $85.70$   &  $84.81$   \\
			\hline
			\multirow{4}{0.3in}{PDA} & SAN~\cite{SAN}  & $80.02$   & $98.64$   &  $100$  &  $81.28$  & $80.58$  &  $83.09$  &  $87.27$   \\
            & IWAN~\cite{IWAN}  & $76.27$   & $98.98$   &  $\mathbf{100}$  &  $78.98$  & $89.46$   &  $81.73$  &  $87.57$  \\
			& PADA~\cite{PADA}  & $86.54$   & $99.32$   &  $\mathbf{100}$  &  $82.17$  & $92.69$   &  $95.41$  &  $92.69$  \\
			& TWINS~\cite{TWINS}  & $86$   & $99.30$   &  $\mathbf{100}$  &  $86.80$  & $94.70$  &  $94.50$  &  $93.60$   \\
			& TSCDA  & $\mathbf{96.84}\pm0.98$   & $\mathbf{100}\pm0.00$   &  $\mathbf{100}\pm0.00$  &  $\mathbf{98.09}\pm0.64$  & $\mathbf{94.75}\pm0.32$ & $\mathbf{96}\pm0.16$  & $\mathbf{97.61}$\\
			\hline 
	\end{tabular}}
\end{table*}

\begin{table*}[htb]
	\caption{Accuracy(\%) on \textit{Office-Home} dataset for PDA tasks.}
	\label{tab:Acc_officehome}
	\centering
    \renewcommand{\tabcolsep}{0.48pc} 
    \renewcommand{\arraystretch}{1.2} 
	\scalebox{0.92}{
		\begin{tabular}{ccccccccccccccc}
			\hline 
			\multicolumn{2}{c}{Method}&  Ar$\rightarrow$Cl  &  Ar$\rightarrow$Pr  & Ar$\rightarrow$Rw  & Cl$\rightarrow$Ar  &  Cl$\rightarrow$Pr & Cl$\rightarrow$Rw & Pr$\rightarrow$Ar & Pr$\rightarrow$Cl & Pr$\rightarrow$Rw & Rw$\rightarrow$Ar & Rw$\rightarrow$Cl & Rw$\rightarrow$Pr & Avg   \\
			\hline
			\multirow{4}{0.3in}{UDA} & ResNet~\cite{ResNet}  & $38.57$   & $60.78$   &  $75.21$  &  $39.94$  & $48.12$  &  $52.90$  &  $49.68$ & $30.91$ & $70.79$ & $65.38$ & $41.79$ & $70.42$ & $53.71$  \\
			& DAN~\cite{DAN}  & $44.36$   & $61.79$   &  $74.49$  &  $41.78$  & $45.21$  &  $54.11$   &  $46.92$ & $38.14$ & $68.42$ & $64.37$ & $45.37$ & $68.85$ & $54.48$ \\
			& DANN~\cite{DANN}  & $44.89$   & $54.06$   &  $68.97$  &  $36.27$  & $34.34$  &  $45.22$   &  $44.08$ & $38.03$ & $68.69$ & $52.98$ & $34.68$ & $46.50$ & $47.39$  \\
			& RTN~\cite{RTN}  & $49.37$   & $64.33$   &  $76.19$  &  $47.56$  & $51.74$  &  $57.67$   &  $50.38$ & $41.45$ & $75.53$ & $70.17$ & $51.82$ & $74.78$ & $59.25$ \\
			\hline
			\multirow{2}{0.3in}{PDA} & PADA~\cite{PADA}  & $51.95$   & $67.00$   &  $78.74$  &  $52.16$  & $53.78$   &  $59.03$  &  $52.61$  & $43.22$ & $78.79$ & $73.73$ & $56.60$ & $77.09$ &$62.06$\\
			& AFN~\cite{AFN}  & $58.93$   & $76.25$   &  $81.42$  &  $70.43$  & $72.97$  &  $77.78$  &  $72.36$  & $55.34$ & $80.40$ & $75.81$ & $60.42$ & $79.92$ & $71.83$\\
			& TSCDA  & $\mathbf{63.64}$   & $\mathbf{82.46}$   &  $\mathbf{89.64}$  &  $\mathbf{73.74}$  & $\mathbf{73.93}$ & $\mathbf{81.43}$  & $\mathbf{75.36}$  & $\mathbf{61.61}$ & $\mathbf{87.87}$ & $\mathbf{83.56}$ & $\mathbf{67.19}$ & $\mathbf{88.80}$ & $\mathbf{77.44}$ \\
			&& $\pm0.06$ & $\pm0.15$ & $\pm0.64$ & $\pm0.61$ & $\pm1.63$ & $\pm0.84$ & $\pm0.56$ & $\pm0.26$ & $\pm0.43$ & $\pm0.47$ & $\pm0.97$ & $\pm0.57$ & \text{--} \\
			\hline 
	\end{tabular}}
\end{table*}

Results on \textit{Office-31} dataset are shown in Table~\ref{tab:Acc_office31}. We observe that the classification results obtained by UDA methods, such as DANN, ADDA and JAN, are worse than directly finetune the ResNet-50 due to negative transfer. TSCDA achieves higher average accuracy ($97.61\%$) than the existing DA methods with large margins ($4\%$) on \textit{Office-31}. In addition, TSCDA outperforms state-of-the-art methods across all the six tasks. In particular, for the more difficult transfer tasks, such as A$\rightarrow$W and A$\rightarrow$D, TSCDA achieves higher accuracy ($96.84\%$ and $98.09\%$) than the state-of-the-art method TWINS with large improvements ($10.8\%$ and $11.3\%$), which further validates the effectiveness of our method.

Results on \textit{Office-Home} benchmark are shown in Table~\ref{tab:Acc_officehome}. Compared with \textit{Office-31}, \textit{Office-Home} has more categories and the four domains are more dissimilar. These can be verified by the classification accuracy of ResNet-50, with an average classification accuracy of only $53.71\%$, which is much lower than the \textit{Office-31} result of $75.64\%$. Thus, \textit{Office-Home} dataset provides a more challenging PDA task. TSCDA achieves higher average accuracy ($77.44\%$) than the existing PDA methods with large margins ($5.6\%$) on \textit{Office-Home}. In particular, TSCDA achieves accuracy of $89.64\%$ and $88.80\%$ on the tasks of Ar$\rightarrow$Rw and Rw$\rightarrow$Pr, and it outperforms the state-of-the-art method AFN by large margins of $8.2\%$ and $8.9\%$ respectively. It can be seen that TSCDA outperforms the existing UDA and PDA methods on all the twelve tasks with larger rooms of improvement. This indicates that our TSCDA can effectively handle the PDA Problems by filtering out the classes in the source-outlier-domain to control negative transfer and learning a target-specific classifier to address label space shift. On the other hand, for the stability of model, the STDs of TSCDA are smaller than $1$ except for Cl$\rightarrow$Pr. In particular, the STDs on Ar$\rightarrow$Cl and Ar$\rightarrow$Pr are near zero, showing the stable performance of TSCDA for repetitiveness.

\begin{table*}[htb]
	\caption{Accuracy (\%) of TSCDA variants on \textit{Office-Home} for PDA tasks.}
	\label{tab:Ablation Study}
	\centering
    \renewcommand{\tabcolsep}{0.49pc} 
    \renewcommand{\arraystretch}{1.2} 
	\scalebox{0.92}{
		\begin{tabular}{cccccccccccccc}
			\hline 
			Method&  Ar$\rightarrow$Cl  &  Ar$\rightarrow$Pr  & Ar$\rightarrow$Rw  & Cl$\rightarrow$Ar  &  Cl$\rightarrow$Pr & Cl$\rightarrow$Rw & Pr$\rightarrow$Ar & Pr$\rightarrow$Cl & Pr$\rightarrow$Rw & Rw$\rightarrow$Ar & Rw$\rightarrow$Cl & Rw$\rightarrow$Pr & Avg   \\
            \hline
			TSCDA-v1 & $56.42$ & $76.47$ & $85.86$ & $65.56$ & $64.43$ & $75.70$ & $65.93$ & $51.40$ & $81.45$ & $75.30$ & $57.73$ & $83.14$ & $69.95$ \\
			TSCDA-v2 & $58.51$ & $81.90$ & $89.18$ & $73.19$ & $71.54$ & $80.45$ & $73.00$ & $55.04$ & $86.20$ & $81.45$ & $60.90$ & $86.67$ & $74.84$ \\
			TSCDA-v3 & $60.18$ & $81.18$ & $89.23$ & $72.91$ & $70.64$ & $77.91$ & $75.11$ & $\mathbf{61.67}$ & $86.53$ & $83.29$ & $66.63$ & $86.27$ & $75.96$ \\
			{\textbf{TSCDA}}  & $\mathbf{63.64}$   & $\mathbf{82.46}$   &  $\mathbf{89.64}$  &  $\mathbf{73.74}$  & $\mathbf{73.93}$ & $\mathbf{81.43}$  & $\mathbf{75.36}$  & $61.61$ & $\mathbf{87.87}$ & $\mathbf{83.56}$ & $\mathbf{67.19}$ & $\mathbf{88.80}$ & $\mathbf{77.44}$ \\
			\hline 
	\end{tabular}}
\end{table*}

\begin{table*}[htb]
\centering
\renewcommand{\tabcolsep}{0.35pc} 
\renewcommand{\arraystretch}{1.2} 
\caption{Accuracy(\%) of TSCDA with different weighting strategies on Office-Home for PDA tasks.}
\label{tab:weighing-strategy}
\begin{tabular}{cccccccccccccc}
\hline
Method      &  Ar$\rightarrow$Cl & Ar$\rightarrow$Pr & Ar$\rightarrow$Rw & Cl$\rightarrow$Ar & Cl$\rightarrow$Pr & Cl$\rightarrow$Rw & Pr$\rightarrow$Ar & Pr$\rightarrow$Cl & Pr$\rightarrow$Rw & Rw$\rightarrow$Ar & Rw$\rightarrow$Cl & Rw$\rightarrow$Pr & Avg\\
\hline
TSCDA(with MMD)     & 63.04 & 72.61 & 87.74 & 70.06 & 69.58 & 78.52 & 68.32 & 61.43 & 84.02 & 82.46 & 59.64 & 86.22 & 73.64\\
TSCDA(with WMMD)      & {\bfseries 64.30} & 79.78 & 88.07 & 72.64 & 71.82 & 79.73 & 73.92 & {\bfseries 63.82} & 86.64 & 83.47 & 63.16 & 86.50 & 76.15\\
TSCDA(with SWMMD)  & 63.64 & {\bfseries 82.46} & {\bfseries 89.64} & {\bfseries 73.74} & {\bfseries 73.93} & {\bfseries 81.43} & {\bfseries 75.36} & 61.61 & {\bfseries 87.87} & {\bfseries 83.56} & {\bfseries 67.19} & {\bfseries 88.80} & {\bfseries 77.44}\\
\hline
\end{tabular}
\end{table*}

\subsection{Analysis}\label{subsec:empirical analysis}

This subsection presents more analysis for TSCDA. Several algorithmic aspects, including ablation study, the weighting strategy, accuracy of pseudo-labeling, parameter sensitivity, feature visualization, sensitivity of the number of target classes and the role of different classifiers are demonstrated as follows.

\begin{figure*}[htb]
	\subfigure[]{\label{fig:target_distribution}
		\begin{minipage}[c]{0.33\textwidth}
			\centering \scalebox{0.33}{
				\includegraphics{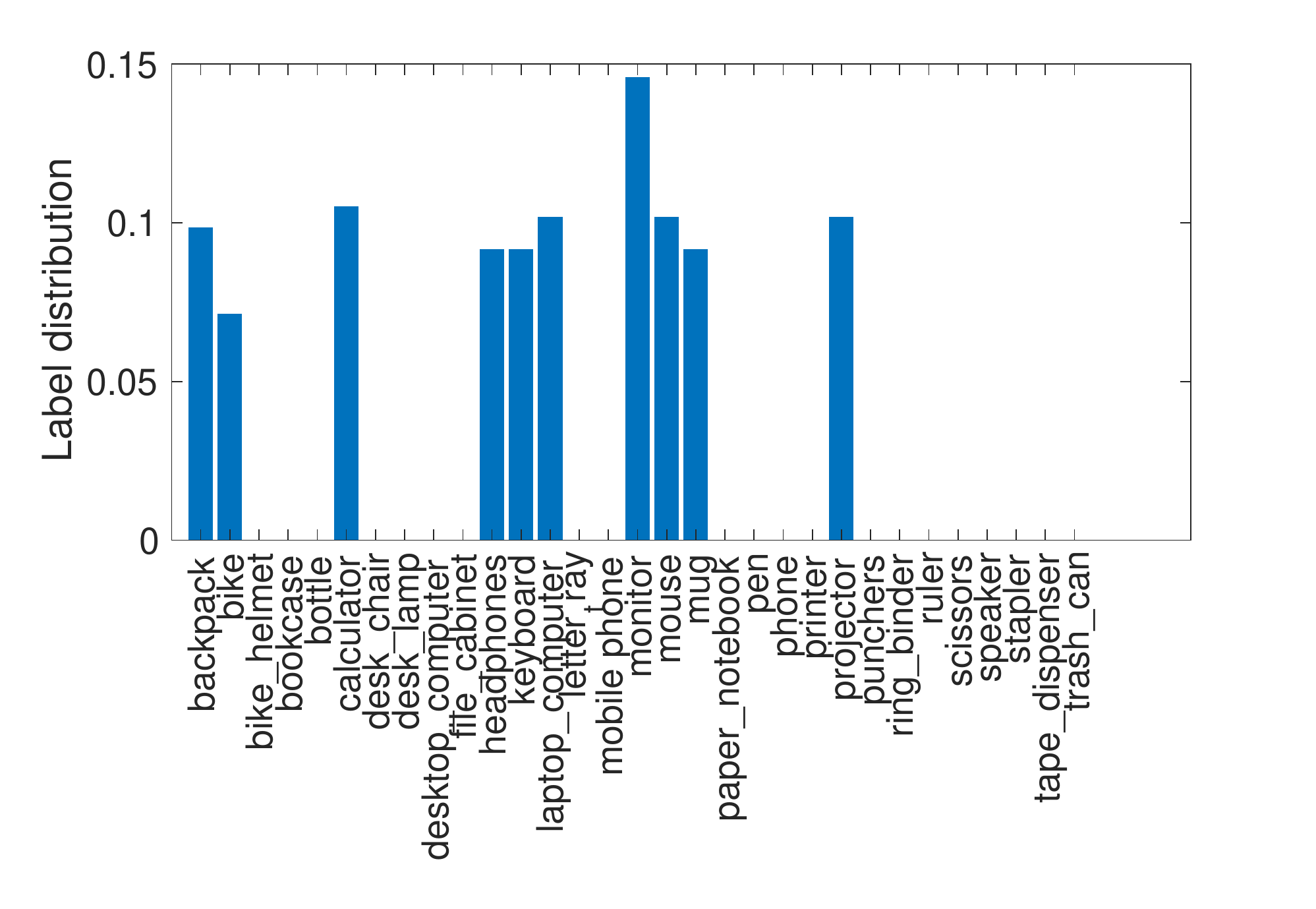}}
	\end{minipage}}
	\subfigure[]{\label{fig:target_flag_distribution}
		\begin{minipage}[c]{0.33\textwidth}
			\centering \scalebox{0.33}{
				\includegraphics{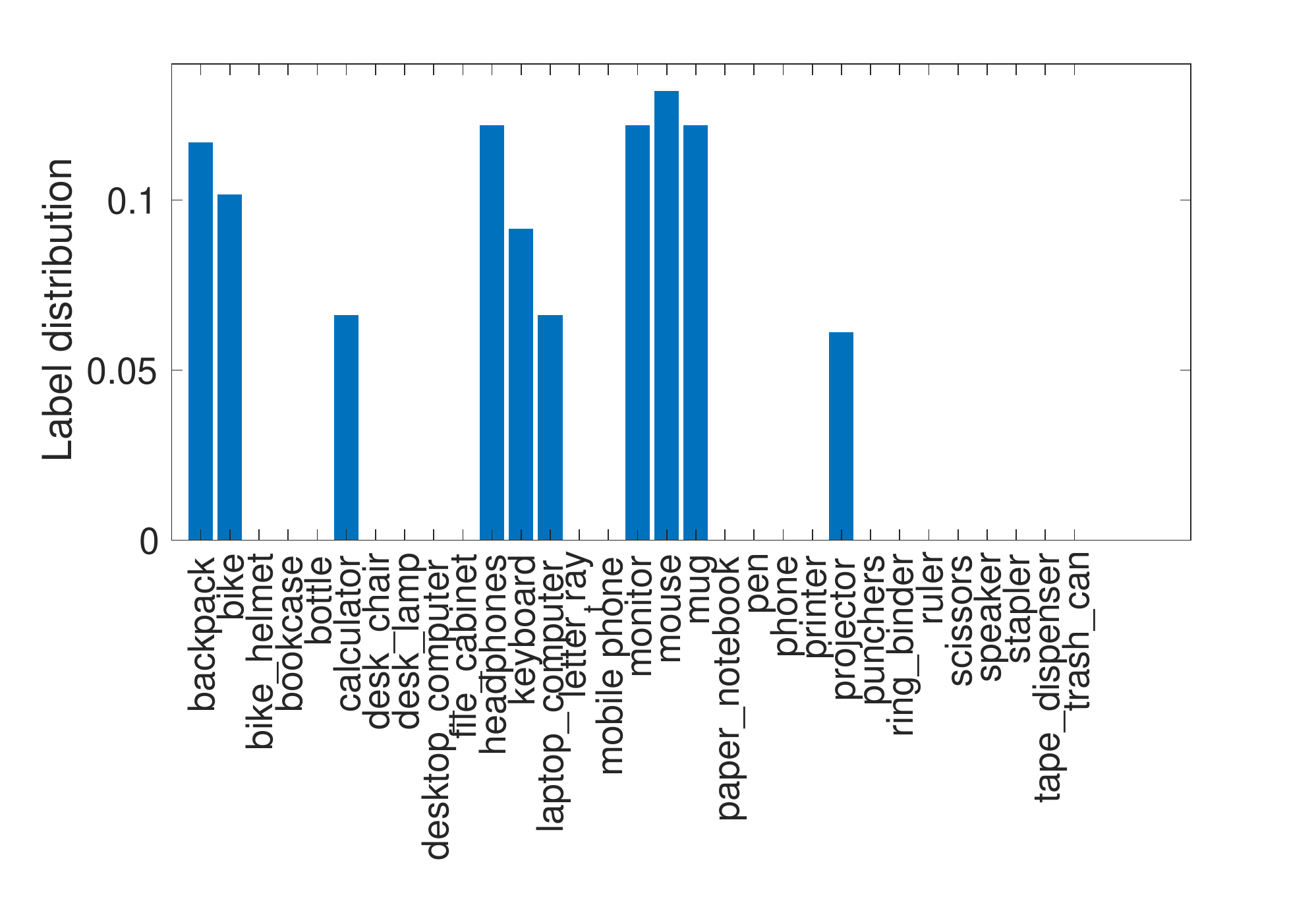}}
	\end{minipage}}
	\subfigure[]{\label{fig:pred_distribution}
		\begin{minipage}[c]{0.33\textwidth}
			\centering \scalebox{0.33}{
				\includegraphics{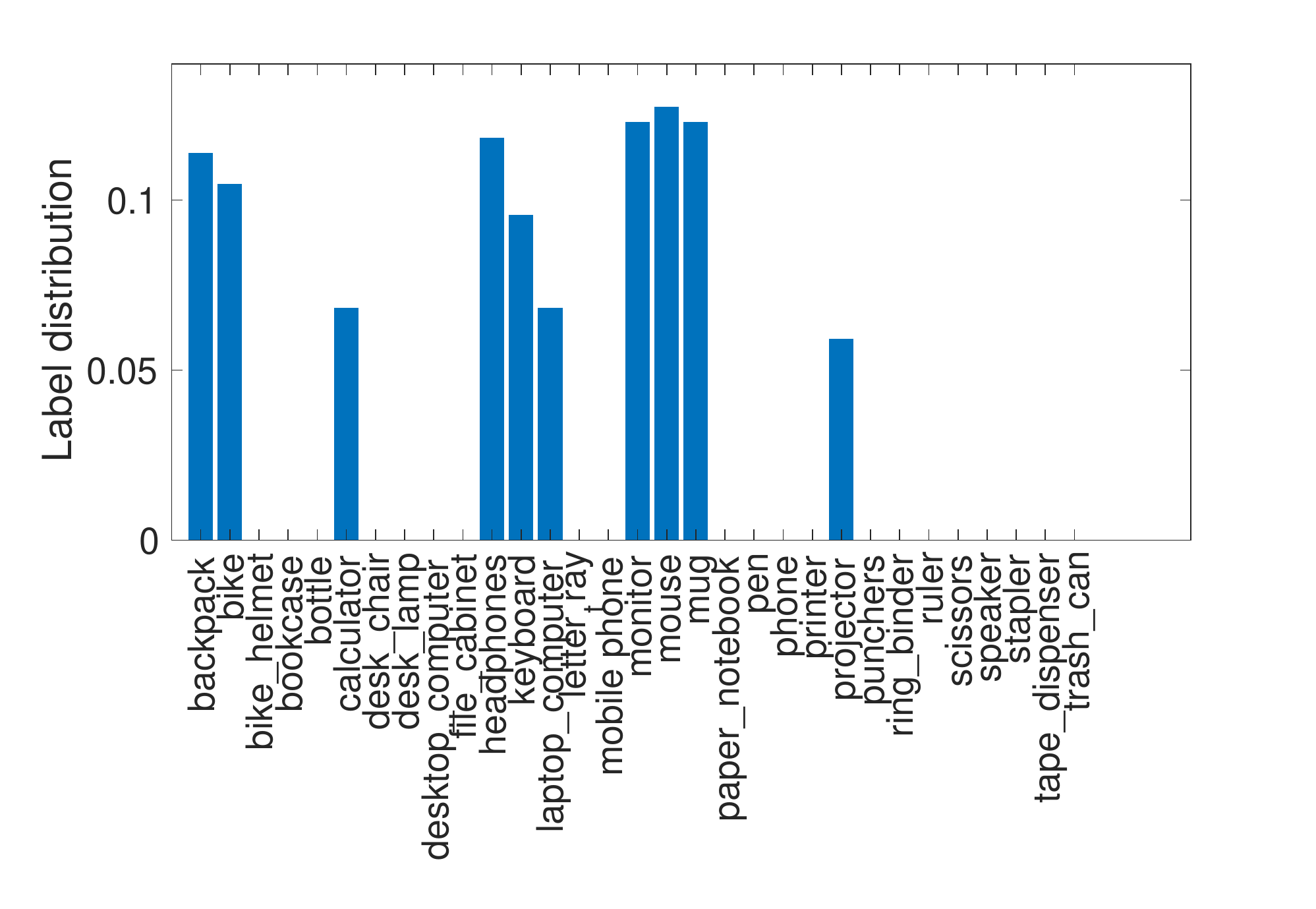}}
	\end{minipage}}
	\caption{Label distribution of target samples. (a) Real label distribution of the whole target domain. (b) Real label distribution of target samples with pseudo-labels. (c) Estimated label distribution of target samples with pseudo-labels.}\label{fig:Label distribution}
\end{figure*}

\subsubsection{Ablation Study}

The loss function of TSCDA contains four different losses, i.e., the classification loss on the source domain $L_{WCE}^s$, the SWMMD loss $L_{SW\!D}$ for partial features alignment, the classification loss $L_{CE}^t$ on target domain, and the inconsistency loss $L_{Con}$. In order to verify the role of target-specific classifier learning module, we perform ablation study with the variant \textbf{TSCDA-v1}, which is built by eliminating from TSCDA the losses $L_{CE}^t$ and $L_{Con}$. To further illustrate the role of PEAL, we perform ablation study with the variant \textbf{TSCDA-v2}, which is built by only eliminating the inconsistency loss $L_{Con}$. To illustrate the role of partial feature alignment, we perform ablation study with the variant \textbf{TSCDA-v3}, which is built by eliminating the loss $L_{SW\!D}$.

Results of ablation study of TSCDA on \textit{Office-Home} are shown in Table~\ref{tab:Ablation Study}. We obtain the following observations: 1) Compared with TSCDA, the performance of TSCDA-v1 drops $7.49\%$. It shows that the target-specific classifier learning module is effective in learning the target-specific classifier and dealing with the classifier shift problem. 2) The classification performance of TSCDA-v2 is higher than TSCDA-v1 and lower than TSCDA. This further indicates the effectiveness of the target-specific classifier learning module and PEAL. 3) TSCDA outperforms TSCDA-v3 on $11$ tasks except for the Pr$\rightarrow$Cl task, which has a little decline ($0.06\%$). It indicates that SWMMD can promote TSCDA to get rid of negative transfer by assigning small weights to source-outlier classes.

\subsubsection{Evaluation of Different Weighting Strategies}

To evaluate the effectiveness of the proposed SWMMD module, we conduct more experiments by replacing SWMMD with MMD and WMMD, respectively, and present the results in Table~\ref{tab:weighing-strategy}.

In all the twelve PDA tasks, the results of MMD are inferior than those of WMMD and SWMMD. Moreover, SWMMD outperforms WMMD on ten of twelve tasks. Finally, the average accuracy of WMMD on these tasks is 76.15\%, while that of SWMMD is 77.44\%. These results validate the effectiveness of SWMMD on differentiating the \textit{source-shared-domain} from the \textit{source-outlier-domain}, which is helpful to addressing the domain alignment and classifier shift problems of PDA tasks.

\subsubsection{Pseudo-Labeling Accuracy}

Since the performance of target-specific classifier relies on the accuracy of pseudo-labels for the target domain, we record the accuracy of pseudo labeling to analyze effectiveness of the PEAL module. Experimental results show that the average classification accuracy of the target domain is up to $98\%$ when $\nu=0.9$. Fig.~\ref{fig:Label distribution} depicts a comparison between the real label distribution and the estimated label distribution for the target samples. Fig.~\ref{fig:target_distribution} and Fig.~\ref{fig:target_flag_distribution} show the real label distributions of the entire target domain and the target sample with pseudo-labels, respectively. Fig.~\ref{fig:pred_distribution} shows the pseudo-labels predicted by the source classifier $C_1$. The comparison of Fig.~\ref{fig:target_distribution} and Fig.~\ref{fig:pred_distribution} shows that pseudo labels predicted by the source classifier cover all source-shared classes and no outliers classes. The comparison of Fig.~\ref{fig:target_flag_distribution} and Fig.~\ref{fig:pred_distribution} shows that the estimated label distribution of the target samples with pseudo labels is highly approximative to the real label distribution. These results show that the target-specific classifier learning can effectively address label space shift. They also show that the target-specific classifier is able to learn a more discriminative decision boundary specific for the target domain with high confidence.

\begin{figure*}[htb]
\subfigure[]{\label{fig:beta_AW}
\begin{minipage}[c]{0.33\textwidth}
\centering \scalebox{0.4}{
\includegraphics{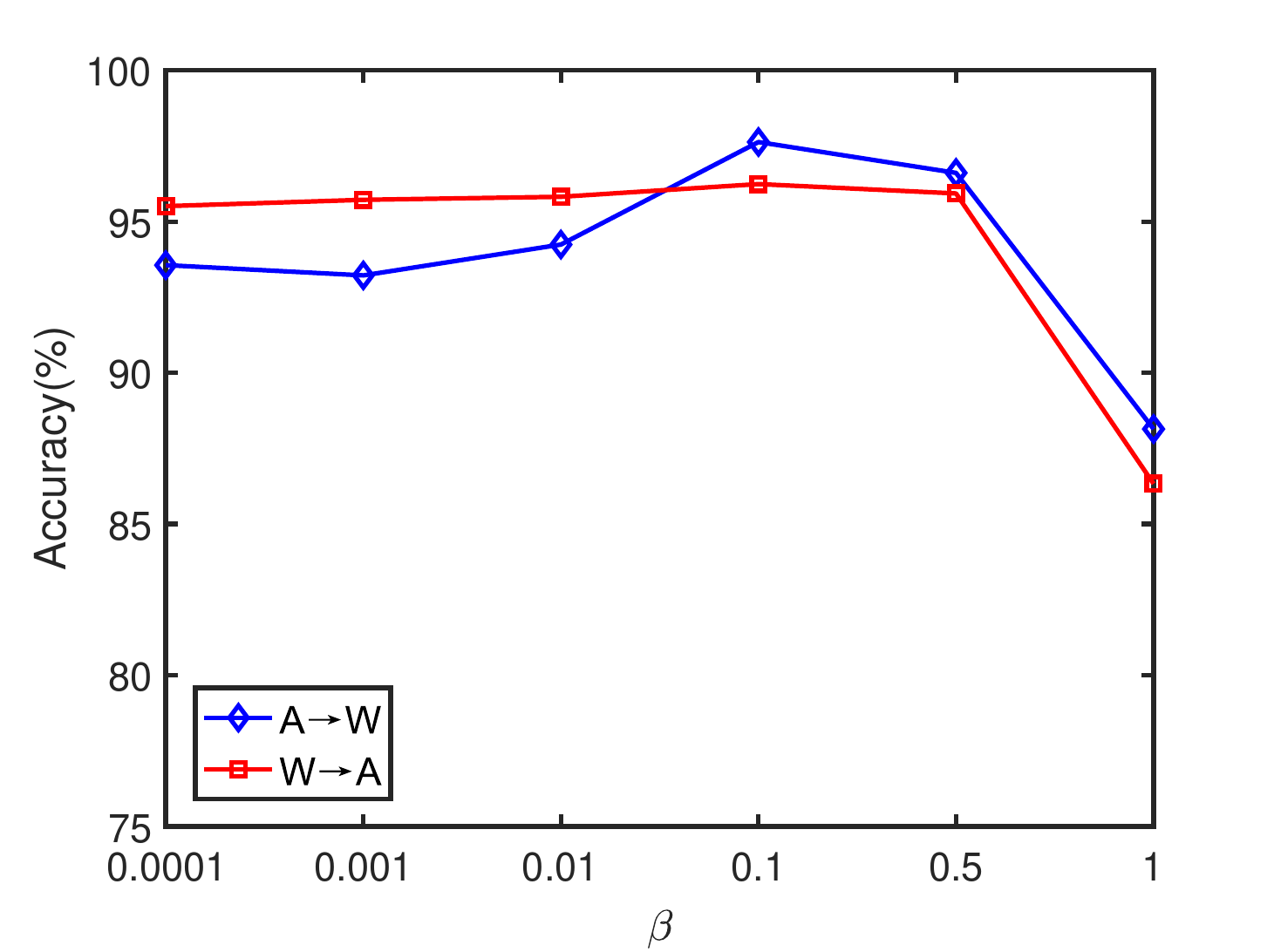}}
\end{minipage}}
\subfigure[]{\label{fig:gamma_AW}
\begin{minipage}[c]{0.33\textwidth}
\centering \scalebox{0.4}{
\includegraphics{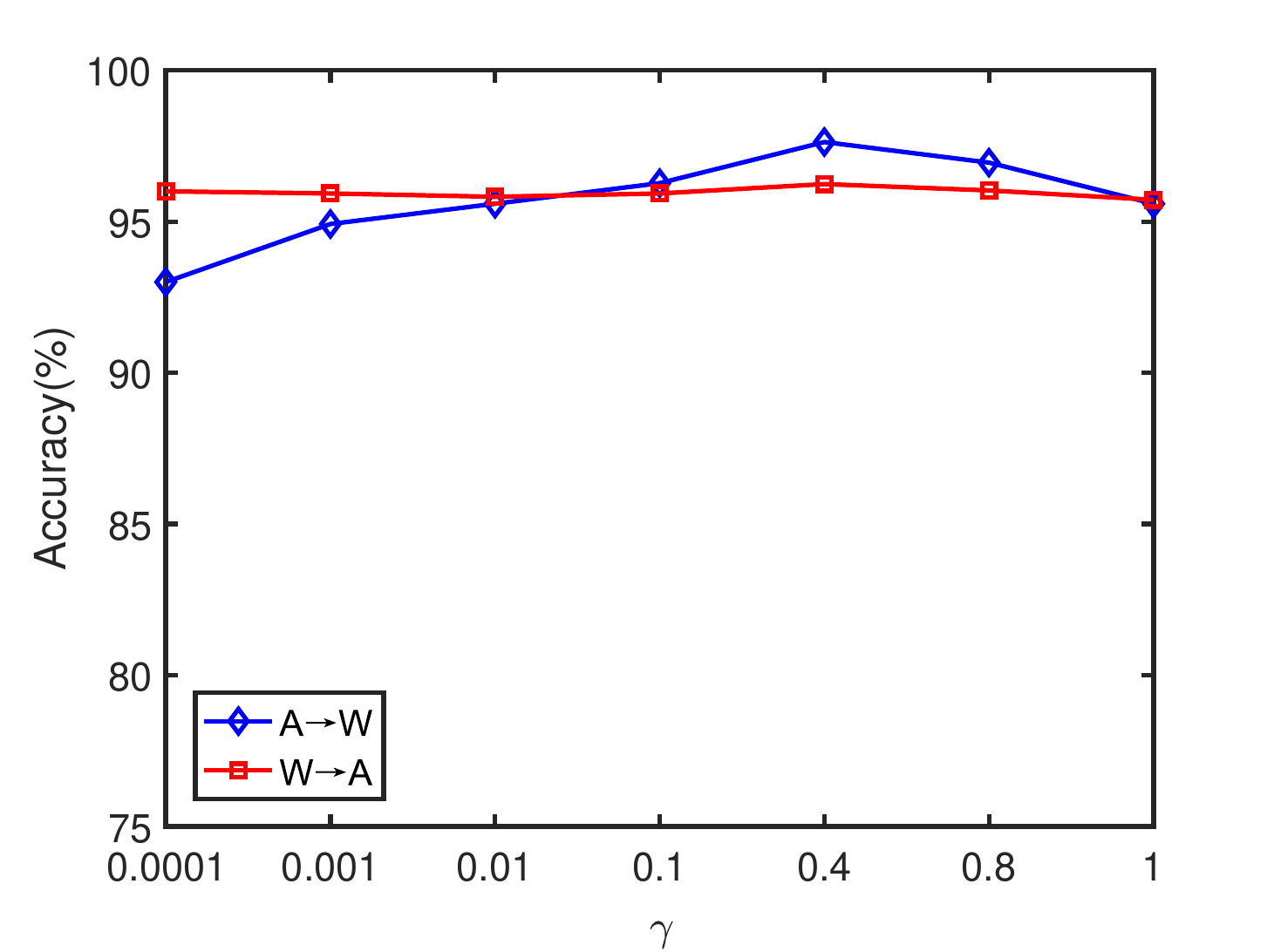}}
\end{minipage}}
\subfigure[]{\label{fig:nu_AW}
\begin{minipage}[c]{0.33\textwidth}
\centering \scalebox{0.4}{
\includegraphics{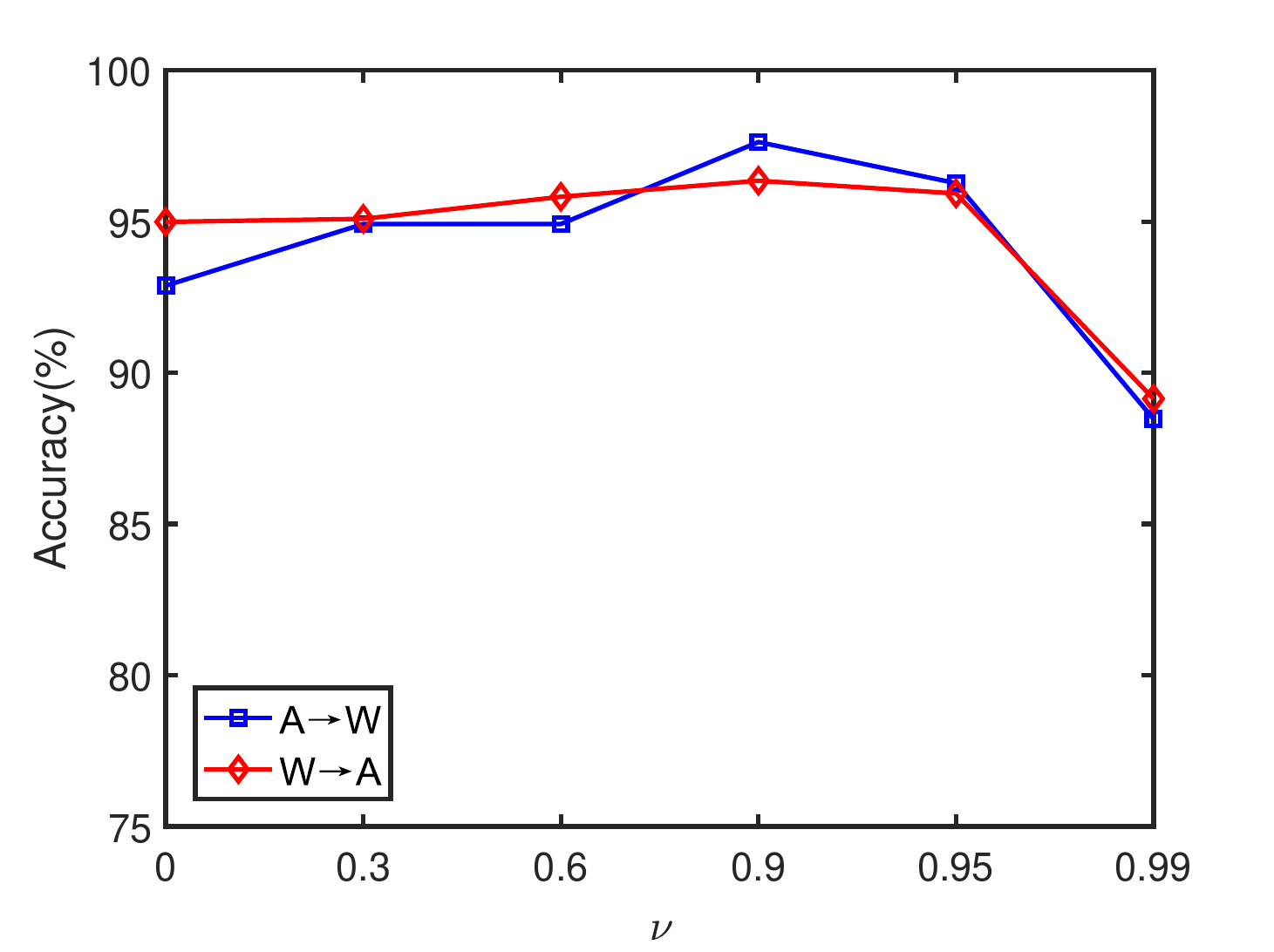}}
\end{minipage}}
\caption{Accuracies of different parameter values. (a) Accuracy w.r.t. $\beta$. (b) Accuracy w.r.t. $\gamma$. (c) Accuracy w.r.t. $\nu$.}\label{fig:parameter sensitivity}
\end{figure*}

\subsubsection{Sensitivity of Parameters}

We now show the empirical analysis results for the sensitivity of hyper-parameters $\beta$, $\gamma$ and $\nu$, by conducting experiments on A$\rightarrow$W and W$\rightarrow$A. Specifically, we use the trial-error approach to obtain the optimal value for each hyper-parameter.

We first fix $\gamma=0.4$, $\nu=0.9$ and vary $\beta$-value within \{1e-4, 1e-3, 1e-2, 1e-1, 5e-1, 1e-0\}. The classification results are shown in Fig~\ref{fig:beta_AW}. Generally, TSCDA obtains the best result around $\beta=0.1$. For task W$\rightarrow$A, the accuracy keeps stable but drops greatly when $\beta=1$. For task A$\rightarrow$W, we can see the accuracy increases as the $\beta$-value becomes larger, but it declines when $\beta$-value exceeds 1e-1. These results also indicate the usefulness of the PEAL module. Accordingly, we use $\beta=0.1$ hereinafter experiments.

The parameter $\gamma$ should be a small value, as a large $\gamma$ is probably to project the source and target data onto the same point with $L_{SW\!D}=0$. We fix $\beta=0.1$, $\nu=0.9$ and vary $\gamma$ within \{1e-4, 1e-3, 1e-2, 1e-1, 4e-1, 8e-1, 1e-0\}. The classification results are shown in Fig.~\ref{fig:gamma_AW}. In the case of A$\rightarrow$W, the accuracy has continuous increment when $\gamma$ changes from 1e-4 to 4e-1. It means that the method pays more attention to the SWMMD term, thus, it shows effectiveness of the partial feature alignment module in PDA. We empirically set $\gamma=0.4$ since TSCDA performs well around it.

The parameter $\nu\in[0,1]$ is used to control the confidence level for pseudo labels, thus, it should be a large value to ensure the accuracy and reliability. We vary it within the range \{0, 0.3, 0.6, 0.9, 0.95, 0.99\}. with $\beta=0.1$ and $\gamma=0.4$. The classification results are shown in Fig.~\ref{fig:nu_AW}. We observe that the accuracy of TSCDA is stable on the W$\rightarrow$A task when $\nu$ is smaller than 0.95, but it degrades rapidly when $\nu$ is very close to 1. For the A$\rightarrow$W task, it obtains the best result at $\nu=0.9$, and the accuracy decrease when larger $\mu$ value is chosen. Therefore, we set $\nu=0.9$ throughout this paper.

\subsubsection{Feature Visualization}

\begin{figure}
\centering
\includegraphics[width=1\columnwidth]{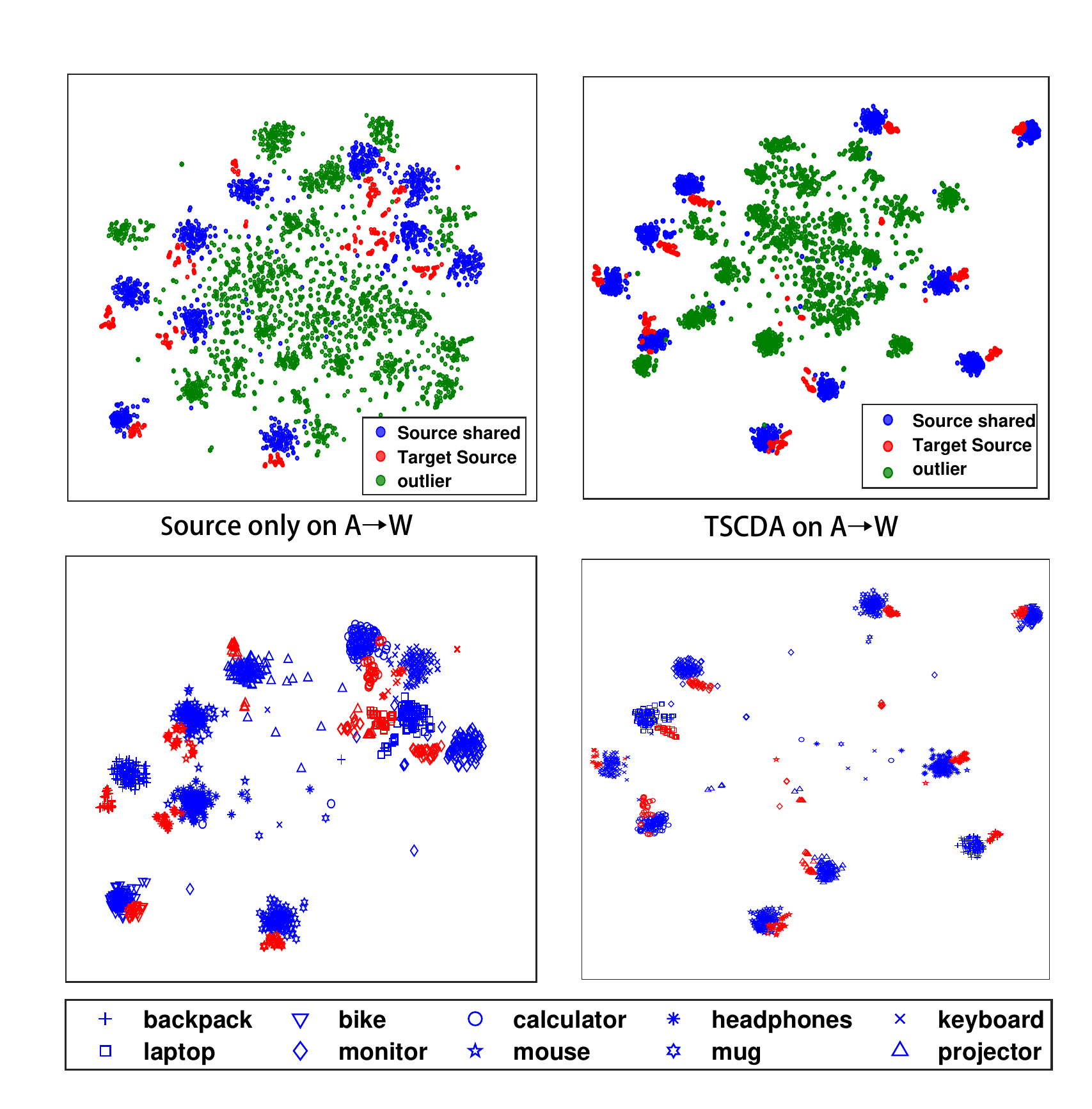}\caption{t-SNE feature visualization. Best viewed in color.}\label{fig:tsne}
\end{figure}

To understand TSCDA better, we present t-SNE feature visualization~\cite{tsne-1} on the pre-adaptive and post-adaptive features of TSCDA, as shown in Fig.~\ref{fig:tsne}.

In particular, to guarantee the class-level alignment could be observed, we use different shapes for samples from different classes and different colors for different domains. The task A$\rightarrow$W with ten shared classes is used for demonstration. The left diagrams show the features obtained by ResNet-50 which is trained only on the source domain, while the right diagrams show the features obtained by TSCDA.

The top line shows the samples annotated in the domain-level, i.e., \textit{source-shared-domain}, \textit{source-outlier-domain} and the target domain. As we can see, features of the source domain and the target domain before adaptation are highly mixed. It is difficult to distinguish samples in the target domain are closer to the \textit{source-shared domain} or the \textit{source-outlier-domain}. As a result, negative transfer can be easily occurred that the target data are mapped to the feature space of the \textit{source-outlier-domain}, and then the domain adaptation performance degenerates. However, in the feature space obtained by TSCDA, we observe that the target features are highly adapted to those of the \textit{source-shared-domain}. It shows that TSCDA can effectively address label space shift to obtain reliable decision boundary for the target domain. In addition, classes in the \textit{source-outlier-domain} are more compact than those before adaptation. It is helpful for us to pay attention to classes appearing in the target domain while neglect the source-outliers.

\begin{figure}[htb]
\subfigure[A$\rightarrow$W]{\label{fig:ClassSensitivity_office31_a2w}
\begin{minipage}[c]{0.23\textwidth}
\centering \scalebox{0.32}{
\includegraphics{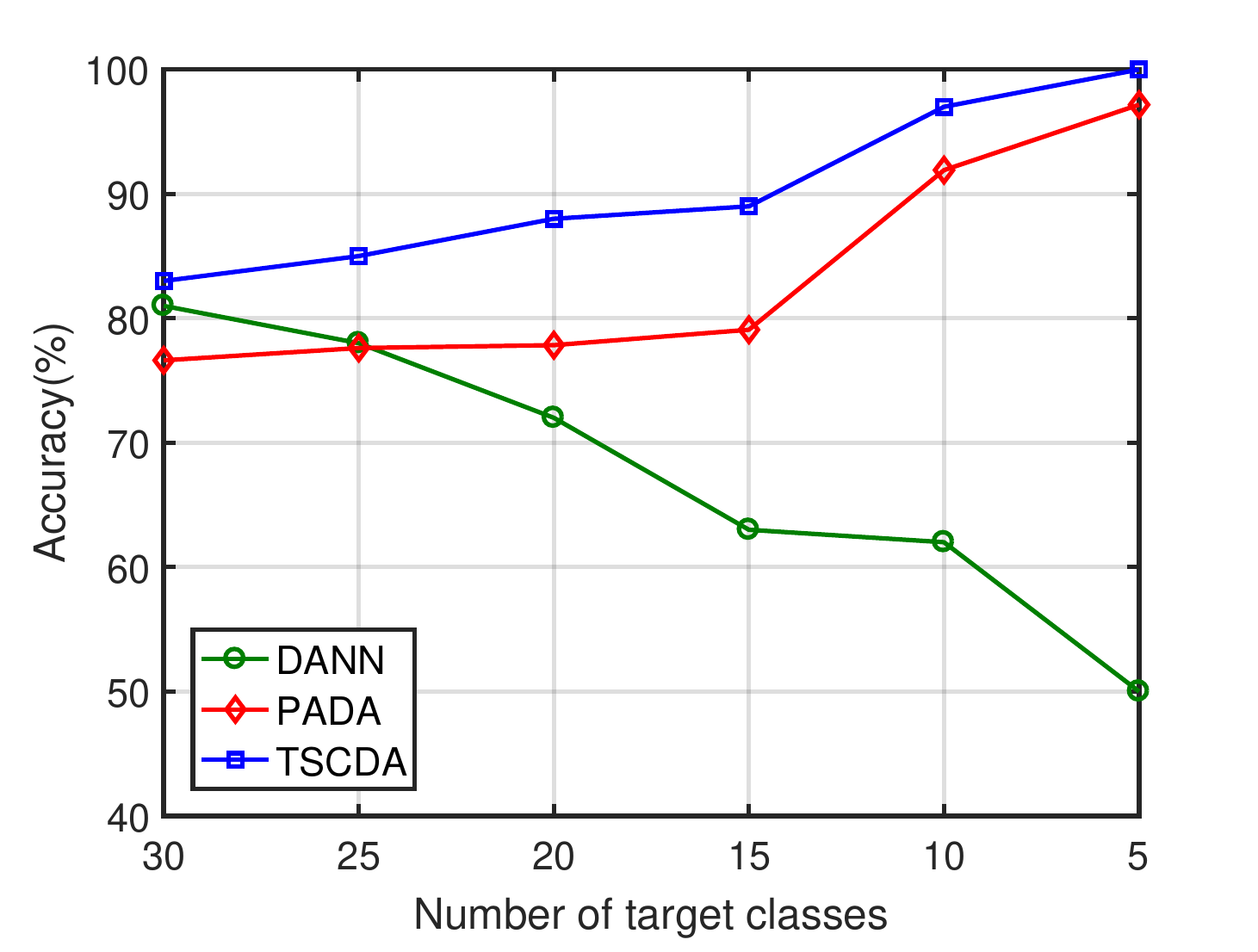}}
\end{minipage}}
\subfigure[W$\rightarrow$A]{\label{fig:ClassSensitivity_office31_w2a}
\begin{minipage}[c]{0.23\textwidth}
\centering \scalebox{0.32}{
\includegraphics{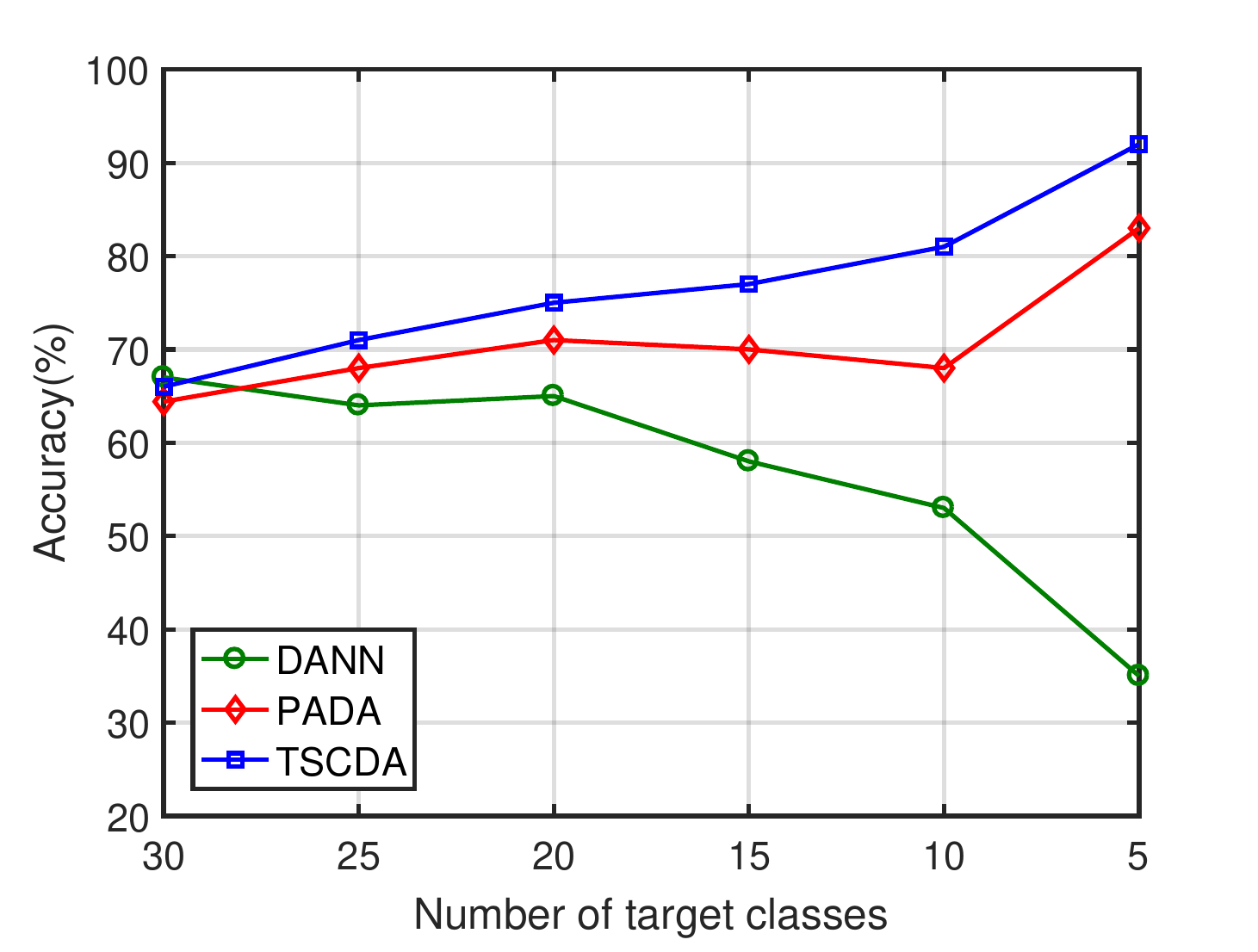}}
\end{minipage}}\\
\subfigure[Ar$\rightarrow$Pr]{\label{fig:ClassSensitivity_officehome_a2p}
\begin{minipage}[c]{0.23\textwidth}
\centering \scalebox{0.32}{
\includegraphics{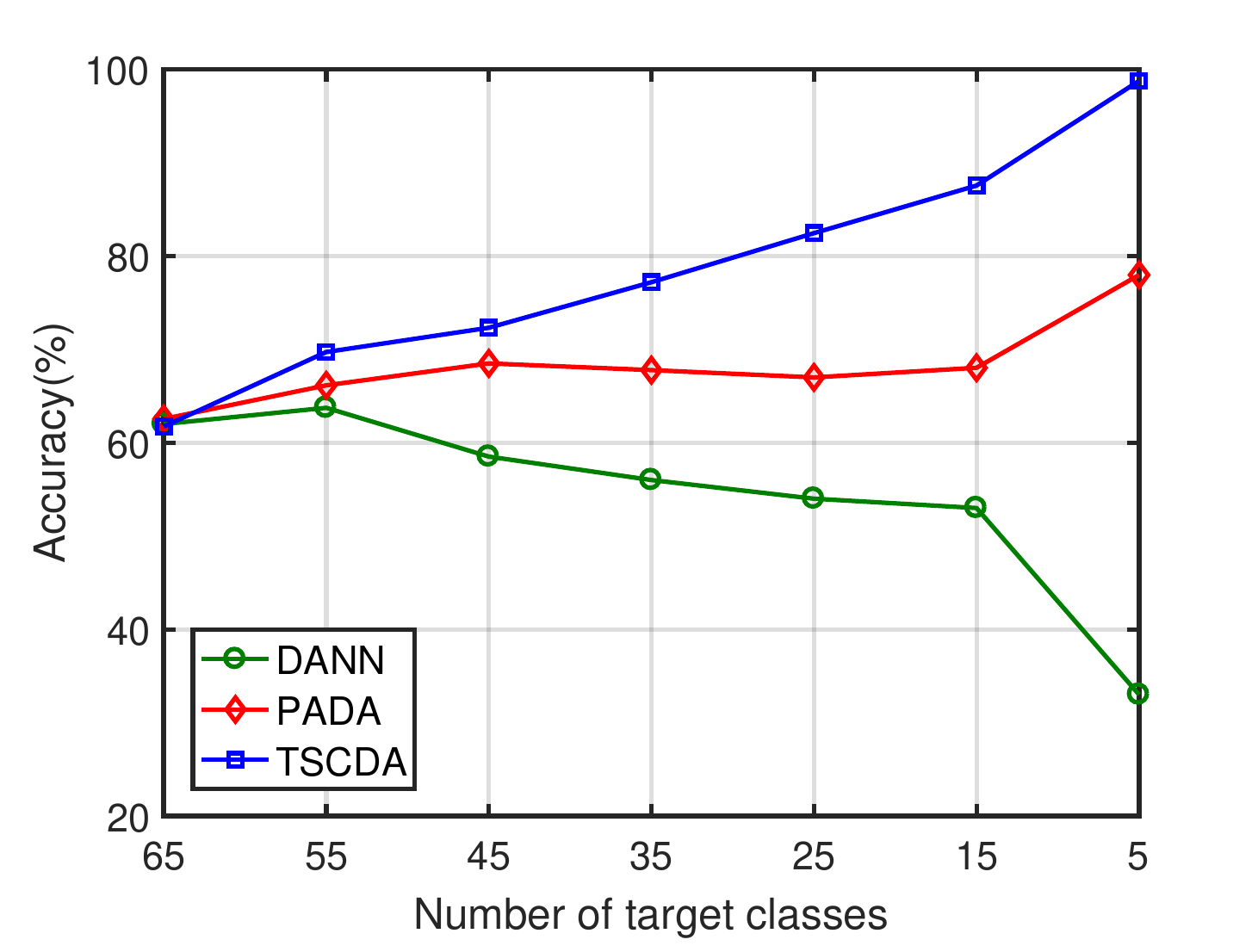}}
\end{minipage}}
\subfigure[Pr$\rightarrow$Ar]{\label{fig:ClassSensitivity_officehome_p2a}
\begin{minipage}[c]{0.23\textwidth}
\centering \scalebox{0.32}{
\includegraphics{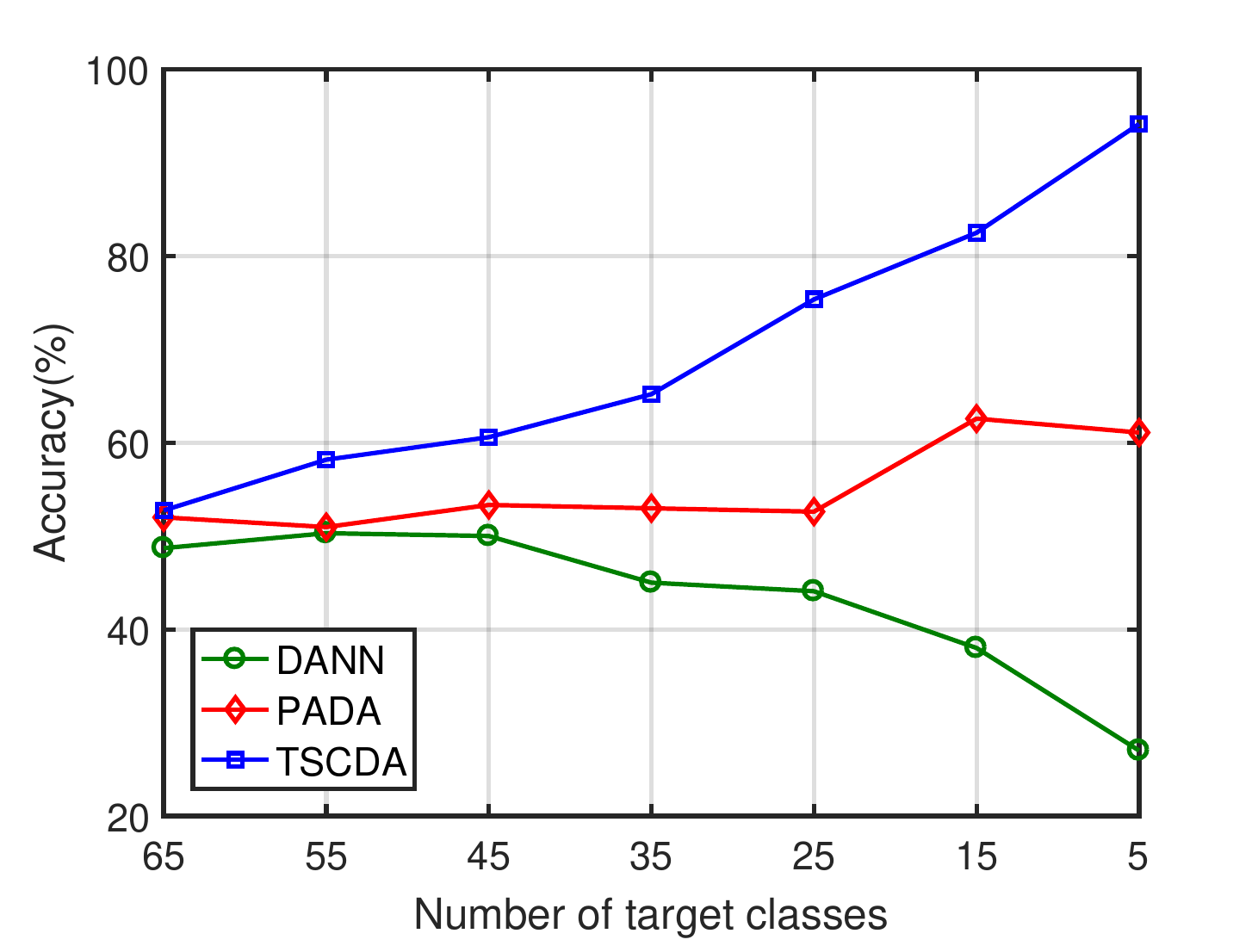}}
\end{minipage}}
\caption{Accuracy with respect to the number of target classes.}\label{fig:ClassSensitivity}
\end{figure}

To present the class-level alignment results, the second line just shows samples in the shared classes. We can see that most samples are aligned correctly, and the margins between different classes are larger than before adaptation. In this perspective, it validates the effectiveness of TSCDA in dealing with the classifier shift and negative transfer problems of PDA tasks.

\subsubsection{Sensitivity of the Number of Target Classes}

We now investigate the performance of TSCDA when the number of target classes changes. DANN~\cite{DANN} and PADA~\cite{PADA} are used as two baselines. The experiments are conducted on \textit{Office-31} and \textit{Office-Home} datasets.

In the experiments of \textit{Office-31}, we fix source domain ($|\mathcal{Y}^s|=31$) and reduce the number of target classes $|\mathcal{Y}^t|$ from 31 to 5. The classification results on two tasks, i.e., A$\rightarrow$W and W$\rightarrow$A, are shown in Fig.~\ref{fig:ClassSensitivity_office31_a2w} and Fig.~\ref{fig:ClassSensitivity_office31_w2a}. In the experiments of \textit{Office-Home}, we also fix source domain ($|\mathcal{Y}^s|=65$) and reduce the number of target classes $|\mathcal{Y}^t|$ from 65 to 5. The classification results on two tasks, i.e., Ar$\rightarrow$Pr and Pr$\rightarrow$Ar, are shown in Fig.~\ref{fig:ClassSensitivity_officehome_a2p} and Fig.~\ref{fig:ClassSensitivity_officehome_p2a}. We can see that the classification accuracy of DANN continuously degrades as the number of target classes decreases, since the size of shared classes becomes small and the classifier shift problem becomes serious. Also note that DANN was just designed for UDA, rather than PDA, thus, it cannot deal with the classifier shift problems well. In contrary, the classification performance of both PADA and TSCDA has a global even continuous improvement as the number of target classes decreases, and TSCDA outperforms PADA on all of these four tasks. In particular, TSCDA has a faster increment in the classification accuracy than PADA. In other words, TSCDA shows remarkable superiority especially when the source label space is much larger than the target label space in PDA tasks.

\subsubsection{The role of different classifiers}

In the TSCDA method, we design three different classifiers, i.e., source classifier $C_1$, target classifier $C_2$ and auxiliary classifier $C_3$. In order to display the role of different classifiers, we evaluate the target domain classification accuracy of different classifiers on four difficult tasks on the Office-31 dataset, i.e., A$\rightarrow$D, A$\rightarrow$D, D$\rightarrow$A and W$\rightarrow$A. Experiment results are shown in Table~\ref{tab:Acc_classifiers}. We observe that the performance of $C_2$ and $C_3$ is significantly better than $C_1$, which verifies our motivation that the source classifier is sub-optimal for the target domain in PDA tasks. At the same time, $C_2$ and $C_3$ achieve similar high classification accuracies, which indicates that PEAL allows the target classifier and auxiliary classifier learn from each other and makes the classifiers more discriminative for the target domain. Both $C_2$ and $C_3$ can be used as the final target domain classifier. In all other experiments in Section~\ref{sec:experiments}, we uniformly report the classification accuracy of $C_2$.

\begin{table}[tb]
	\caption{Accuracy(\%) of different classifiers on Office-31 dataset}
	\label{tab:Acc_classifiers}
	\centering
    \renewcommand{\tabcolsep}{1pc} 
    \renewcommand{\arraystretch}{1} 
		\begin{tabular}{ccccc}
			\hline 
			Classifier &   A$\rightarrow$W  & A$\rightarrow$D  &  D$\rightarrow$A & W$\rightarrow$A  \\
			\hline
			$C_1$  & $89.15$   & $92.36$   &  $91.54$  & $93.84$\\
			$C_2$  & $96.95$   & $97.45$   &  $94.78$  &  $95.82$  \\
			$C_3$  & $96.95$   & $98.09$   &  $94.36$  &  $96.03$ \\
			\hline 
	\end{tabular}
\end{table}

\begin{figure}[htb]
\subfigure[]{\label{fig:c3-a2w}
\begin{minipage}[c]{0.23\textwidth}
\centering \scalebox{0.31}{
\includegraphics{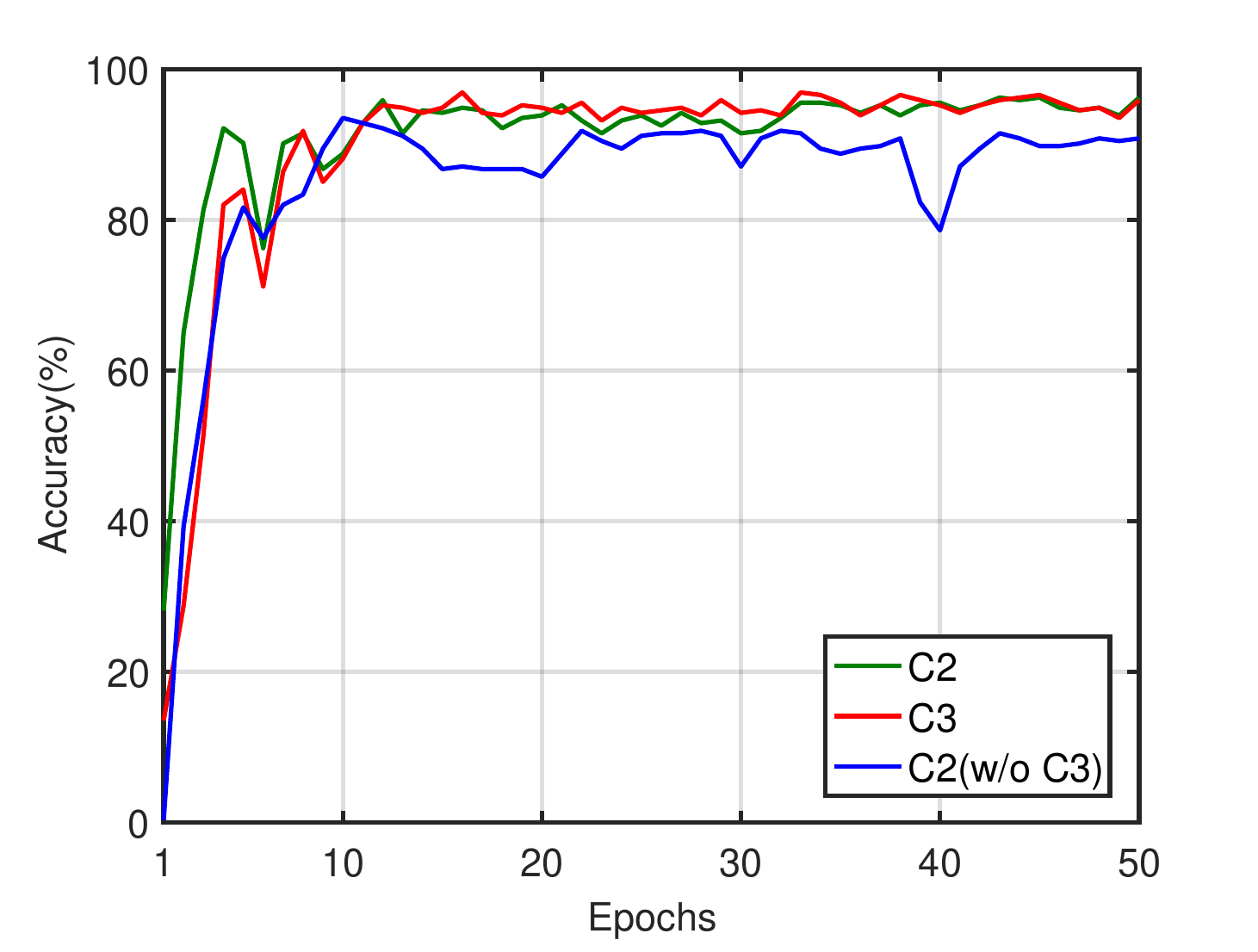}}
\end{minipage}}
\subfigure[]{\label{fig:c3-ar2pr}
\begin{minipage}[c]{0.23\textwidth}
\centering \scalebox{0.365}{
\includegraphics{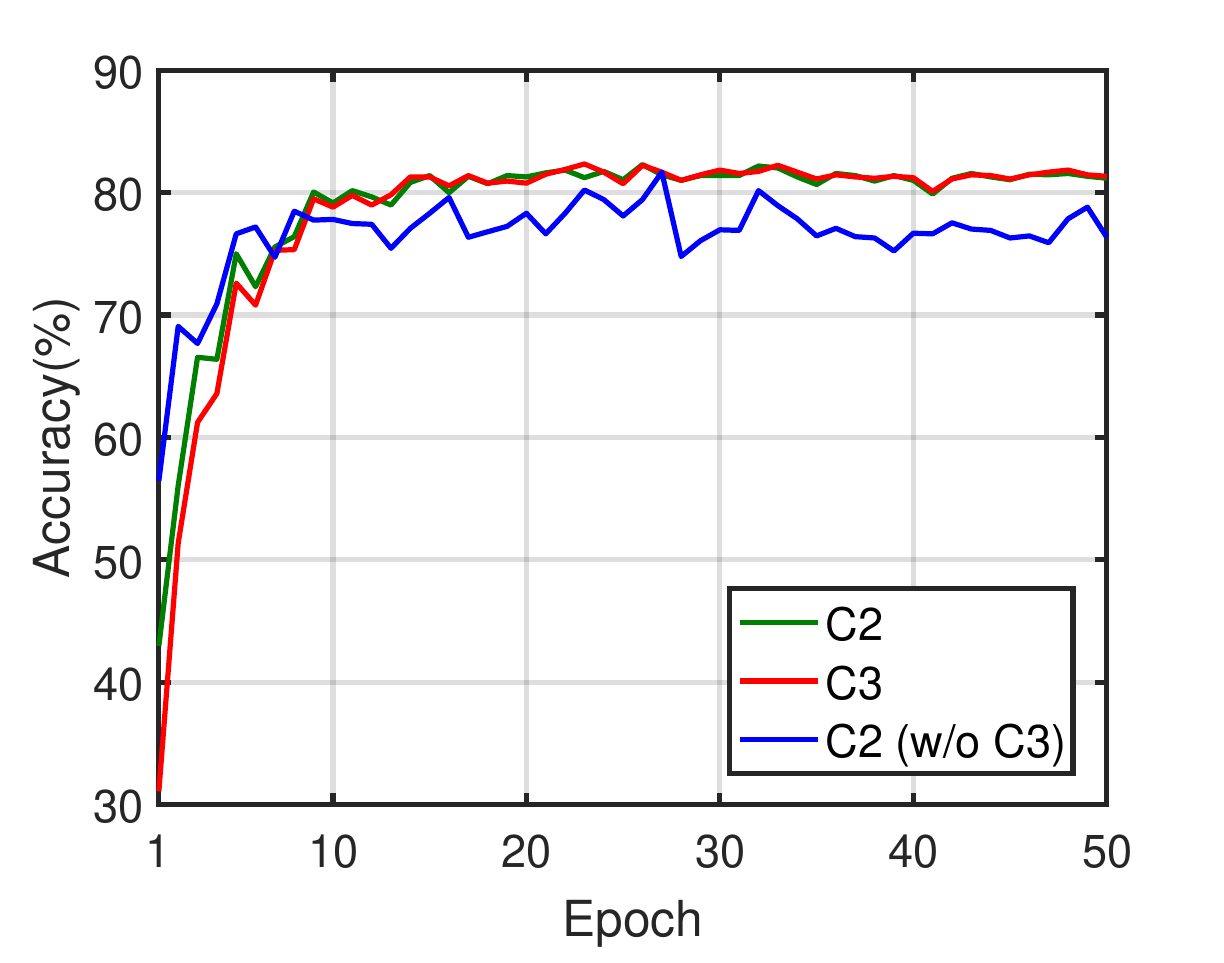}}
\end{minipage}}
\caption{The role of $C_3$ in classification performance. (a) A$\rightarrow$W. (b) Ar$\rightarrow$Pr.}\label{fig:role-c3}
\end{figure}

We further conduct more experiments to show the performance of $C_2$ and $C_3$ during and after learning. Fig.~\ref{fig:role-c3} presents the classification accuracies with different training strategies, namely, one learns with both $C_2$ and $C_3$, while the other learns with $C_2$ only. Fig.~\ref{fig:c3-a2w} shows the results on task A$\rightarrow$W and Fig.~\ref{fig:c3-ar2pr} on task Ar$\rightarrow$Pr. When $C_3$ is used in the PEAL module, we can see that the results of $C_2$ are higher than those of $C_3$ at the beginning of the iterations, but soon they start to approach each other and converge to the same classification accuracies. In contrary, when $C_3$ is removed from PEAL, the accuracies of $C_2$ which are shown by ``$C_2$(w/o $C_3$)'' in the figures cannot achieve the high level obtained by the other manner. Therefore, the role of $C_3$ in improving the final classification performance is validated.

\section{Conclusion}\label{sec:conclusion}		

This paper presents a novel method to tackle PDA problems. Previous methods do not consider the essential classifier shift scenario, and they just share and use the source classifier to test directly samples in the target domain. These are obviously sub-optimal for PDA tasks. The proposed TSCDA can not only deal with the negative transfer between the \textit{source-outlier-domain} and the target domain by partial feature alignment with SWMMD, but also address the classifier shift problem by learning a target-specific classifier. In particular, the PEAL module forces features to distribute compact in the shared feature space, thus, it is helpful to learn a more discriminative decision boundary for the target domain. Comprehensive experiments and comparisons show that TSCDA achieves state-of-the-arts performance for PDA problems.

How to extend the PEAL module to more extensive learning scenes, and formulate solid theoretical foundations, are our future works.

\bibliographystyle{IEEEtran}
\bibliography{CAN}	

\end{document}